\documentclass[runningheads]{llncs}

\usepackage{eccv}

\usepackage{eccvabbrv}

\usepackage{graphicx}
\usepackage{booktabs}

\usepackage[accsupp]{axessibility}  

\usepackage{hyperref}

\usepackage{orcidlink}
\usepackage{wrapfig}
\usepackage{multirow}
\usepackage{booktabs} 
\usepackage{makecell} 
\usepackage{array}
\usepackage{subcaption}
\usepackage{float}

\usepackage[dvipsnames]{xcolor}

\usepackage{pifont}       
\usepackage{bbding}       
\usepackage{fontawesome}  
\usepackage{enumitem} 

\usepackage{wrapfig}

\usepackage{etoc}
\usepackage{footmisc}
\usepackage[misc]{ifsym}

\makeatletter
\def\blfootnote{\gdef\@thefnmark{}\@footnotetext}
\makeatother

\begin{document}

\title{Towards In-Context Tone Style Transfer with a Large-Scale Triplet Dataset} 

\titlerunning{In-Context Tone Style Transfer with a Large-Scale Triplet Dataset}

\author{
Yuhai Deng\inst{1, 2}$^*$
\and
Huimin She\inst{4}$^*$
\and
Wei Shen\inst{4}$^\dag$
\and 
Meng Li\inst{4} 
\and 
Ruoxi Wu\inst{4} 
\and 
Lunxi Yuan\inst{4} 
\and 
Xiang Li\inst{2, 1, 3}$^\textrm{\Letter}$
}

\authorrunning{Y.~Deng et al.}

\institute{VCIP, School of Computer Science, Nankai University  
\and 
Nankai International Advanced Research Institute (Shenzhen Futian)
\and
AAIS, Nankai University \\
\and OPPO AI Center, OPPO Inc.\\
\email{yhdeng.me@gmail.com, xiang.li.implus@nankai.edu.cn
}}
\maketitle
\blfootnote{*: Equal Contribution. \dag: Project Leader.  \textrm{\Letter} : Corresponding Author.}

\begin{abstract}
Tone style transfer for photo retouching aims to adapt the stylistic tone of the reference image to a given content image. 
However, the lack of high-quality large-scale triplet datasets with stylized ground truth forces existing methods to rely on self-supervised or proxy objectives, which limits model capability. To mitigate this gap, we design a data construction pipeline to build TST100K, a large-scale dataset of 100,000 content-reference-stylized triplets. At the core of this pipeline, we train a tone style scorer to ensure strict stylistic consistency for each triplet.
In addition, existing methods typically extract content and reference features independently and then fuse them in a decoder, which may cause semantic loss and lead to inappropriate color transfer and degraded visual aesthetics.
Instead, we propose ICTone, a diffusion-based framework that performs tone transfer in an in-context manner by jointly conditioning on both images, leveraging the semantic priors of generative models for semantic-aware transfer. 
Reward feedback learning using the tone style scorer is further incorporated to improve stylistic fidelity and visual quality.
Experiments demonstrate the effectiveness of TST100K, and ICTone achieves state-of-the-art performance on both quantitative metrics and human evaluations. 
{\hypersetup{pdfborder={0 0 0}}
The \href{https://dengyuhai.github.io/ICTone_Project/}{\textcolor{magenta}{project page}} is available online.
}
\end{abstract}

\section{Introduction}
\begin{figure*}[tb]
  \centering
  \includegraphics[width=1.0\linewidth]{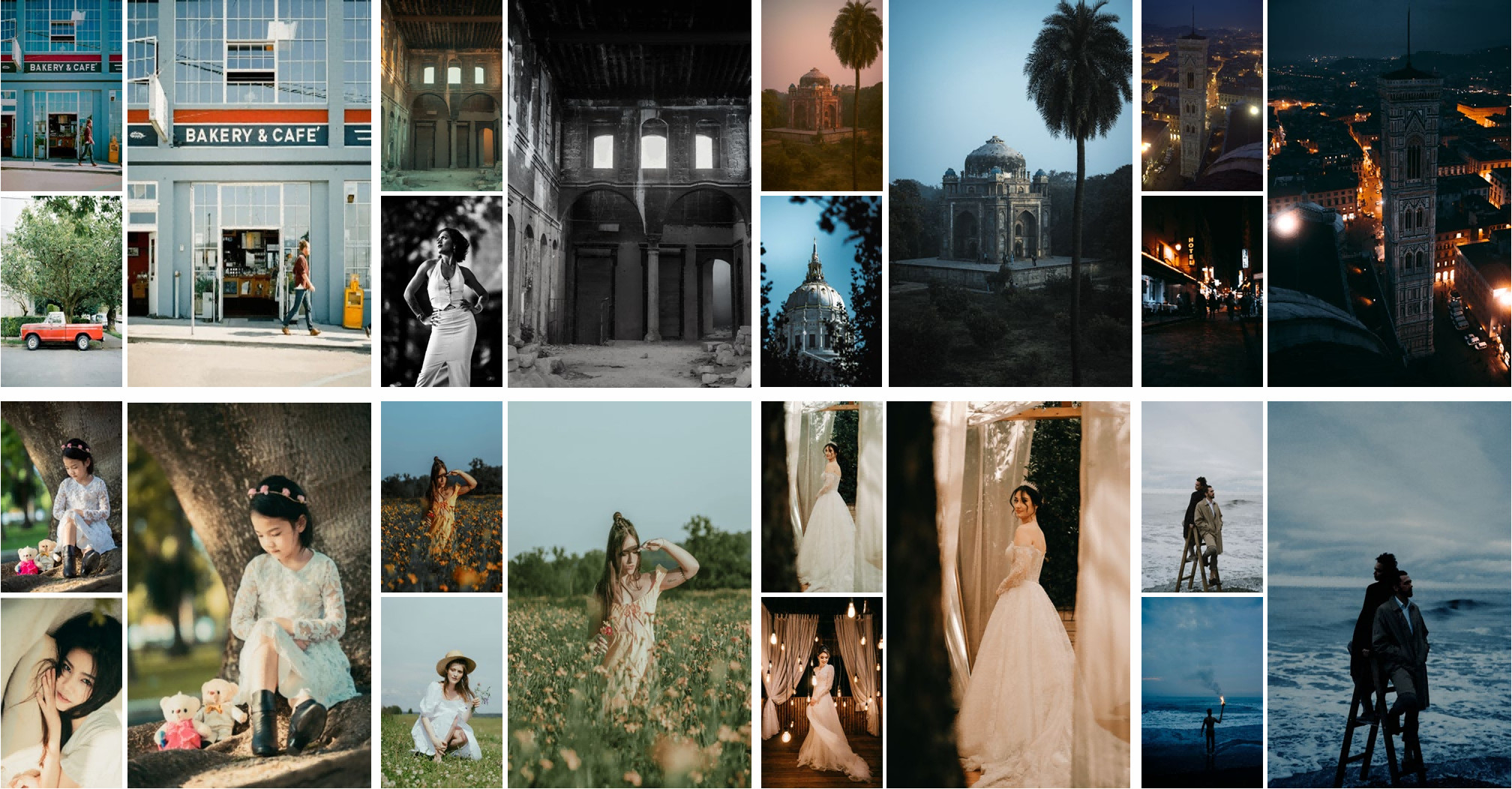}
  \caption{Showcases of our method performing tone style transfer across diverse scenarios.}
  \label{fig:teaser}
\end{figure*}
Photo retouching and color manipulation are fundamental to computational photography and image editing, allowing images to be adapted to diverse photographic styles. Among these techniques, tone style transfer~\cite{ke2023neural,lv2024color} aims to adapt the photographic appearance of a content image according to a reference image. As illustrated in Fig.~\ref{fig:teaser}, it transfers tone-related attributes such as color distribution, luminance, contrast, and saturation while preserving the semantic content and structural integrity of the content image. Compared with traditional color transfer, which mainly matches global color statistics, and artistic style transfer, which often alters textures and structures, tone style transfer operates at the level of photographic aesthetics and requires semantically aware adaptation across different image regions.

The absence of ground-truth stylized targets for content-reference pairs fundamentally limits the learning capability of existing tone style transfer methods~\cite{lv2024color}. Consequently, these methods must rely on self-supervised training~\cite{ke2023neural} or proxy style losses~\cite{wen2023cap,yoo2019photorealistic,li2018closed,ho2021deep}, which provide only indirect supervision for learning reference-conditioned tone transfer. For example, Neural Preset~\cite{ke2023neural} constructs self-supervised training pairs by applying different presets to the same content image, where one preset-retouched image is used as the reference and the other serves as the pseudo target. Nevertheless, as shown in Tab.~\ref{tab:sota_cmp}, training Neural Preset with ground-truth content–reference–stylized triplets significantly improves its performance, demonstrating that self-supervision inherently constrains model capability.

To address the lack of high-quality paired data, we propose a data construction pipeline to build a large-scale dataset of content-reference-stylized triplets. The key challenge in constructing such a dataset is the absence of a reliable metric for tone style similarity. Without such a metric, dataset construction faces an inherent trade-off between quality and scalability: manual retouching produces accurate pairs but is prohibitively expensive, while applying predefined presets is scalable but inevitably introduces noisy and perceptually inconsistent triplets. Specifically, the same preset may produce unreliable tonal correspondences when applied to images with diverse content. To overcome this dilemma, we develop a dedicated tone style scorer through a two-stage training paradigm. Specifically, the scorer is first trained with weakly supervised contrastive learning~\cite{khosla2020supervised} on noisy preset-generated pairs to learn general tone style similarity, and subsequently fine-tuned on human-ranked pairs for improved perceptual alignment. By leveraging this robust scorer, we can effectively ensure strict stylistic consistency between the reference and stylized images for each preset-generated triplet. In addition, an aesthetic model~\cite{sheng2023aesclip} is also incorporated to filter out low-quality images and improve the overall visual quality of the dataset.

While large-scale datasets enhance model performance, existing methods still exhibit inappropriate color transfer. This issue fundamentally arises from their architectural design, as most models utilize independent encoders to extract features from the content and reference images separately before fusing them within a decoder. This process inevitably leads to semantic loss, severely restricting the overall semantic perception capability of the model. For instance, as shown in Fig.~\ref{fig:vis_compare}, the retrained CAP-VSTNet incorrectly maps object colors onto human faces, resulting in unnatural skin coloration.  
To resolve such semantic misalignments, we formulate tone style transfer as an in-context generation task~\cite{wu2025less,zhang2025context,li2025ic} and introduce ICTone. As a diffusion-based framework, ICTone treats the content and reference images as a joint contextual input, leveraging the strong semantic priors of large-scale generative models~\cite{rombach2022high,esser2024scaling,wu2026representation,duan2025diffretouch,chang2026pertouch} to achieve accurate, structure-aware tone style transfer. To further improve stylistic fidelity and aesthetic appeal, we incorporate reward feedback learning~\cite{xu2023imagereward} using the trained tone style scorer as the reward signal.

Our main contributions are as follows:

\begin{itemize}[leftmargin=0.5cm]
    \item We construct \textbf{TST100K}, the first large-scale dataset of content-reference-stylized triplets, providing reliable ground truth for training robust tone style transfer models. Furthermore, we introduce \textbf{TST2K}, a carefully curated high-quality benchmark of 2,000 triplets that enables precise evaluation.

    \item We develop a \textbf{tone style scorer} to solve a long-standing challenge in the field: how to effectively measure the stylistic tone similarity between two images. This scorer serves a dual role by ensuring strict stylistic and perceptual consistency during data construction, while providing a reward signal during model training to align outputs with perceptual style fidelity.
    
    \item We propose \textbf{ICTone}, a diffusion-based framework that performs tone transfer as an in-context generation task and incorporates reward feedback learning to enhance stylistic alignment and aesthetic quality. Extensive quantitative and human evaluations demonstrate that ICTone achieves state-of-the-art performance, successfully validating the effectiveness of both our framework and the TST100K dataset.
\end{itemize}

\section{Related Work}

\subsection{Datasets for Photo Retouching}
Existing photo retouching datasets mainly focus on global tone and color enhancement. Representative datasets include the MIT-Adobe FiveK dataset~\cite{bychkovsky2011learning} and PPR10K~\cite{liang2021ppr10k}, which provide large-scale input–output pairs for supervised enhancement. However, these datasets are designed for single-image enhancement under fixed expert styles, without modeling content–style correspondences or reference-driven transfer. Recent large-scale triplet datasets for artistic style transfer, such as OmniStyle~\cite{wang2025omnistyle}, provide over one million content–style–stylized triplets across diverse artistic styles. However, they focus on artistic transformations that allow texture and structural changes, making them unsuitable for photorealistic tone style transfer, which requires strict preservation of scene geometry and pixel-level details. In contrast, datasets specifically designed for photorealistic style transfer remain limited. 
DPST~\cite{luan2017deep} provides photograph pairs of content and style images but lacks ground-truth stylized outputs.
In contrast, PST50~\cite{gong2025sa} includes ground-truth results, but contains only 50 pairs, limiting diversity and benchmarking scale for tone style transfer models.
To address these limitations, we introduce two large-scale tone style transfer datasets: TST100K, a high-quality dataset that enables supervised training, and TST2K, a carefully curated benchmark for evaluation.

\subsection{Style Transfer} 
Tone style transfer originates from the broader field of image style transfer, which aims to apply the visual style of a reference image to a content image. Style transfer is commonly divided into artistic style transfer~\cite{gatys2016image,liu2025sad} and photorealistic style transfer~\cite{yoo2019photorealistic}.
Early artistic style transfer methods~\cite{gatys2016image,johnson2016perceptual,li2017universal,an2020ultrafast} represent style using Gram matrices and perform stylization through iterative optimization, successfully reproducing artistic textures but often introducing artifacts and structural distortions.
Early approaches~\cite{reinhard2002color,pitie2005n,pitie2007linear,xiao2006color} to photorealistic style transfer focus on matching low-level color statistics between images, aligning color histograms or per-channel means and variances between the content and reference images in carefully designed color spaces~\cite{marschner2021fundamentals,gevers2012color}, achieving basic color and tone alignment.
More recent methods adopt learning-based frameworks to improve transfer quality while preserving structural fidelity~\cite{yoo2019photorealistic,ke2023neural,liu2023universal,larchenko2025color,wu2024goal,chiu2022photowct2}. For instance, PhotoWCT~\cite{li2018closed,wen2023cap,an2021artflow} performs whitening–coloring transforms followed by edge-preserving smoothing to maintain spatial consistency. More recently, SA-LUT~\cite{gong2025sa} employs spatially adaptive 4D look-up tables to achieve real-time, high-quality photorealistic tone transfer with improved spatial fidelity.
However, these methods often exhibit limited semantic understanding, leading to suboptimal performance in cross-scene transfer. To address this limitation, we propose ICTone, which leverages the rich priors of generative diffusion models to enable more accurate and robust semantic-aware transfer.

\section{Method}
In this section, we first present our dataset construction, including image collection, tone style scorer training, and the overall pipeline, and then introduce ICTone, highlighting its in-context formulation and reward feedback learning.

\subsection{Dataset Construction}
\begin{figure*}[tb]
  \centering
  \includegraphics[width=1.0\linewidth]{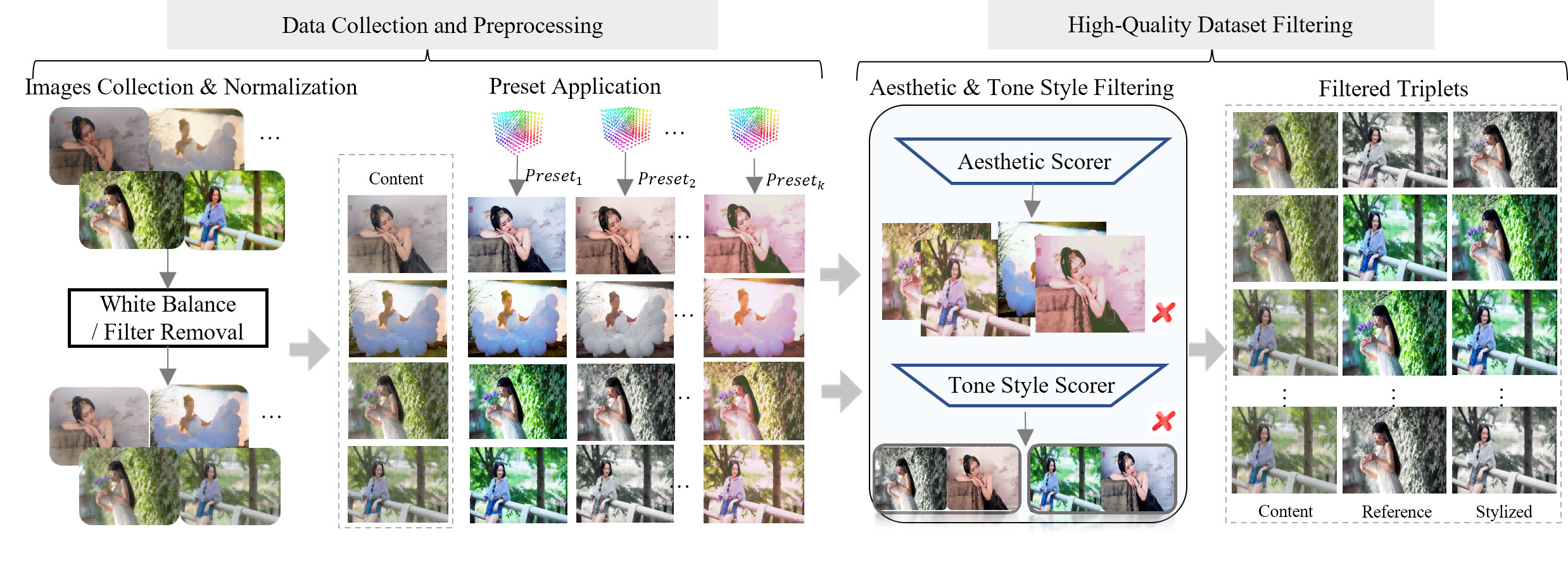}
  \caption{
  Overview of the dataset construction pipeline. Left: data collection and preprocessing, where images undergo white balance correction and filter removal, followed by applying diverse tone presets to generate stylized candidates. Right: high-quality filtering using an aesthetic scorer and a tone style scorer to select pairs with high stylistic consistency and visual quality, forming the final content-reference-stylized triplets.}
  \label{fig:DataPipeline}
\end{figure*}

Existing tone style transfer methods often rely on self-supervised training due to the lack of high-quality paired data, limiting their ability to learn accurate tone styles. Paired datasets can be created either by manual retouching, which is accurate but costly, or by applying presets, which is scalable but often inconsistent across diverse content. To address this, we train a tone style scorer to evaluate tone similarity and use it to select high-quality preset-generated pairs, which is the core of our dataset construction pipeline. As illustrated in Fig.~\ref{fig:DataPipeline}, our pipeline includes image collection, normalization, and high-quality filtering, producing a large-scale set of content–reference–stylized triplets with consistent tone and high visual quality.

\subsubsection{Data Collection and Preprocessing}
\label{ssbsec:collection}
\mbox{}\\
The training data for the tone style scorer is collected from multiple public datasets, including MIT-Adobe FiveK~\cite{bychkovsky2011learning}, PPR10K~\cite{liang2021ppr10k}, COCO~\cite{lin2014microsoft}, Food-101~\cite{bossard2014food}, and Landscape HQ~\cite{skorokhodov2021aligning}. These datasets cover a diverse range of visual domains, such as portraits, landscapes, and urban scenes.
For constructing TST100K, we primarily utilize images from PPR10K and MIT-Adobe FiveK due to their high image quality. 
Presets are collected from Adobe Lightroom and professional photography communities, where they are pre-categorized into scenarios such as portrait, landscape, night, lifestyle, and food. For each category, redundant presets are removed using LAB histogram filtering and manual inspection, resulting in approximately 3,000 distinct presets.
Before applying presets, all images are normalized to mitigate pre-existing tonal biases. Specifically, automated white balance correction~\cite{afifi2020deep} and filter removal~\cite{yeo2022cair} are performed. We then apply curated presets to normalized images, creating a substantial pool of different images sharing the same tone style. 
\subsubsection{Tone Scorer Training}
\begin{figure*}[tb]
	\centering
	\includegraphics[width=1.0\linewidth]{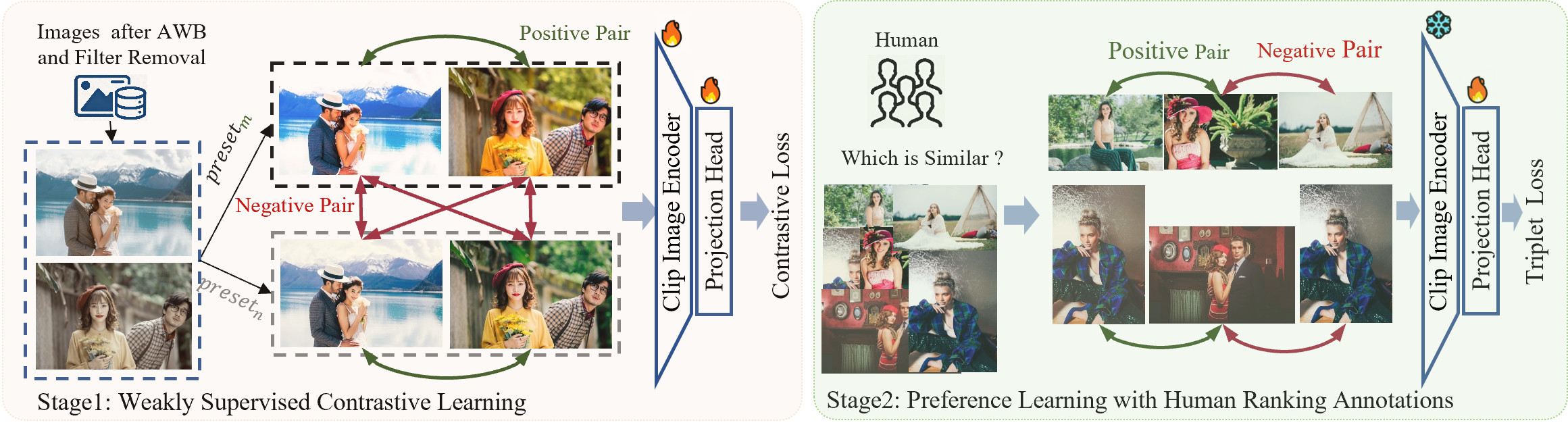}
	\caption{Overview of the two-stage tone style scorer training pipeline. It combines weakly supervised contrastive learning and preference learning for tone style alignment.}
	\label{fig:discriminator_pipeline}
\end{figure*}

\mbox{}\\
To ensure tone consistency during dataset construction, we train a tone style scorer to evaluate the tonal similarity between image pairs. We adopt a CLIP architecture~\cite{radford2021learning} for the tone style scorer, leveraging its strong capability in learning discriminative visual embeddings~\cite{zhang2026crystal,li2026wowseg,zhang2025unichange}. The image encoder follows the ViT-B/16 backbone, and the projection head maps image features into a normalized tone embedding space. These embeddings are later used to measure tone style similarity when constructing the paired dataset. 
As illustrated in Fig.~\ref{fig:discriminator_pipeline}, we train the tone scorer in two stages to achieve reliable tone similarity estimation. In the first stage, we employ weakly-supervised contrastive learning using preset-generated data, where image pairs produced by the same preset are treated as positive samples, while those from different presets are treated as negative samples. This stage enables the model to capture coarse tone representations from large-scale data. In the second stage, we fine-tune the model with a small set of human-annotated ranking data to refine its perception of tonal similarity and obtain the final tone style scorer for dataset construction. 

\textbf{Weakly-supervised contrastive learning using presets.} In this stage, images processed with the same preset are regarded as positives, while those processed with different presets serve as negatives. This formulation enables the model to learn tone similarity based on preset labels without requiring manual annotations. The training objective is formulated as a supervised contrastive loss~\cite{khosla2020supervised}, defined as:
\begin{equation}
	\mathcal{L}_{Con}
	= -\frac{1}{N} \sum_{i=1}^{N}
	\frac{1}{|P(i)|} \sum_{p \in P(i)}
	\log \frac{\exp\!\left( \mathbf{z}_{i} \cdot \mathbf{z}_{p} / \tau \right)}
	{\sum_{a \ne i} \exp\!\left( \mathbf{z}_{i} \cdot \mathbf{z}_{a} / \tau \right)},
    \label{equ:WSL}
\end{equation}
where $N$ denotes the batch size, $P(i)$ is the set of positives sharing the same preset label with sample $i$, $\mathbf{z}_i \in \mathbb{R}^d$ is the normalized feature embedding of sample $i$, and $\tau$ is a temperature parameter. This formulation guides the model to capture discriminative tone representations that encode global color distributions and illumination patterns. To ensure generalization across diverse tone variations, we train the scorer on the previously collected dataset (Section~\ref{ssbsec:collection}). Furthermore, we apply image blurring as a data augmentation strategy to encourage the model to capture tone-related characteristics.

\textbf{Preference learning with human-ranking annotations.} 
While contrastive learning on preset-generated pairs offers discriminative supervision, it may not fully match human perceptual assessments. In practice, a single preset often produces perceptually divergent results across images, leading to noisy positive pairs. 
We therefore employ multiple discriminators with diverse backbones (VGG, ResNet, and ViT) to identify hard samples with inconsistent predictions, which are then refined by human ranking, resulting in a curated set of 20K human-ranking annotations. Each annotation contains several candidates with similar content but different tone adjustments. Human evaluators rank these candidates based on tone consistency and visual harmony.
We adopt a triplet loss to refine the tone style scorer. The loss $\mathcal{L}_{\text{Tri}}$ is defined as follows:
\begin{equation}
	\mathcal{L}_{\text{Tri}} = \max\left\{0,\; d(\mathbf{z}_{a},\mathbf{z}_{p}) - d(\mathbf{z}_{a},\mathbf{z}_{n}) + m\right\},
    \label{equ:tri}
\end{equation}

Here, $d(u,v)$ is the cosine distance, defined as: $d(u, v) = 1 - \frac{\langle u, v \rangle}{\lVert u \rVert_2 \, \lVert v \rVert_2}$, $\mathbf{z}_{a}$ denotes the feature embedding of anchor image, $\mathbf{z}_{p}$ denotes feature embedding of a positive sample ranked higher or perceptually similar, and $\mathbf{z}_{n}$ denotes feature embedding of a negative sample ranked lower or perceptually dissimilar. The loss encourages the anchor to be closer to the positive than to the negative by at least a margin $m$.
This stage is essential for the scorer to evaluate tone alignment in accordance with human-perceived preferences.

\subsubsection{Triplet Dataset Construction}
\mbox{}\\
We construct a large-scale dataset of content-reference-transferred triplets, where the content image represents an unedited photo, the reference provides the target tonal appearance, and the transferred image serves as the retouched result. 

\textbf{Reference-stylized pair verification.}
To ensure high visual quality and reliable tone alignment, we employ a dual-constraint filtering strategy during dataset construction. Specifically, an aesthetic non-degradation constraint and a tone consistency constraint are applied as complementary filtering criteria. For aesthetic quality control, we adopt an aesthetic assessment model~\cite{sheng2023aesclip} to prevent visual degradation introduced by preset-based stylization. Given an original image and its stylized counterpart, we retain only samples whose aesthetic score is no lower than that of the original image prior to preset application. 
In addition, we introduce a tone style scorer to evaluate tone alignment between the reference and stylized images by measuring cosine similarity in a learned tone embedding space. Reference–stylized pairs with similarity below 0.8 are discarded to enforce tone consistency and ensure reliable training supervision. This dual verification process ensures the stylized image is both visually appealing and tone-consistent with its reference, forming reliable supervision signals for training.

\textbf{Content diversity augmentation.}
To improve robustness and better reflect diverse real-world inputs, additional content variations are introduced during training. Specifically, transformations such as color desaturation, exposure reduction, and LUT-based color perturbations are randomly applied to the content image. These operations are performed online and preserve the underlying semantic structure while modifying illumination, contrast, and color distribution. This augmentation strategy expands the appearance diversity of the content domain, enabling the model to maintain stable tone transfer performance across varied input conditions.

Finally, we construct TST100K, a large-scale dataset of 100,000 content-reference-stylized triplets with consistent tone style alignment. Based on TST100K, we further curate a representative benchmark subset termed TST2K. Each triplet is manually reviewed for tone alignment and visual quality. The subset contains 2,000 curated triplets, comprising samples from TST100K and manually retouched image pairs. As illustrated in Fig.~\ref{fig:dataset_overview}, TST2K spans diverse real-world scenarios, including portrait, food, landscape, night, and lifestyle scenes, providing a comprehensive evaluation setting.

\subsection{In-Context Tone Style Transfer}
\begin{figure}[t]
  \centering

  \begin{minipage}[t]{0.55\textwidth}
    \centering
    \includegraphics[width=\linewidth]{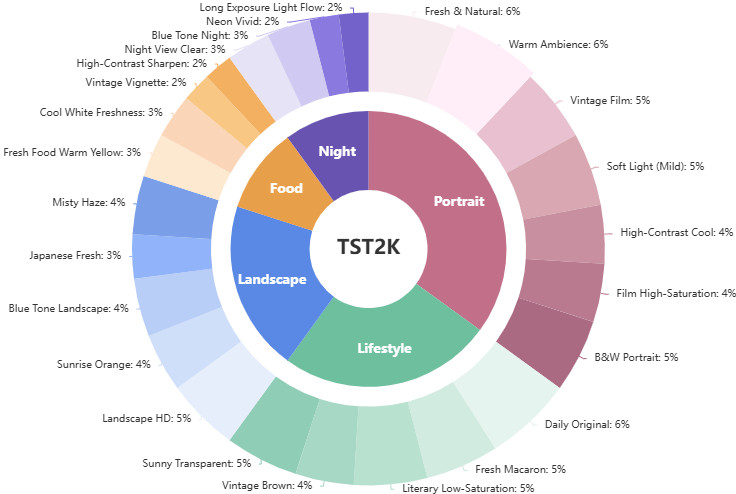}
    \caption{The distribution of TST2K benchmark.}
    \label{fig:dataset_overview}
  \end{minipage}
  \hfill
  \begin{minipage}[t]{0.43\textwidth}
    \centering
    \includegraphics[width=\linewidth]{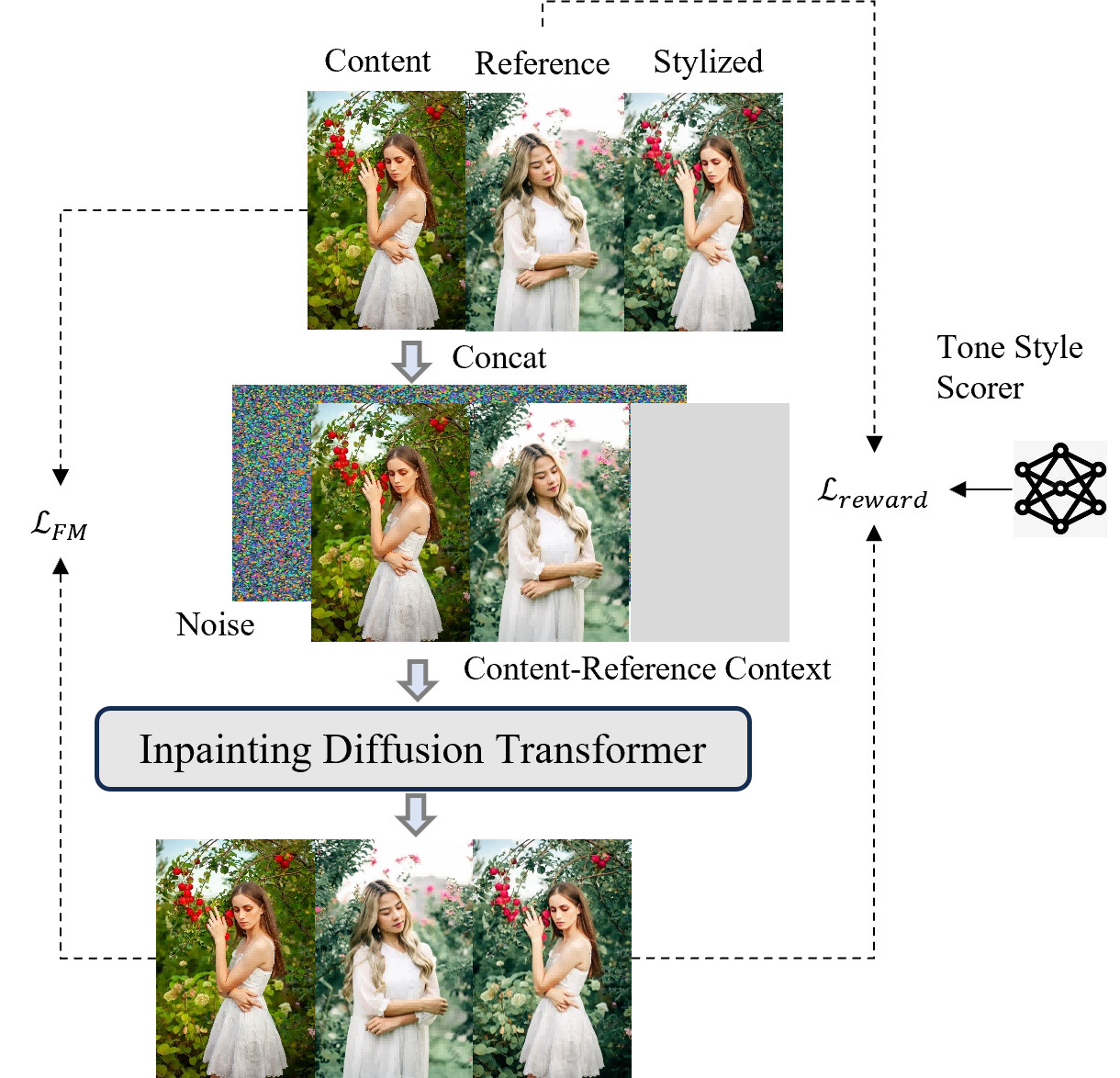}
    \caption{Overview of the in-context model training pipeline.}
    \label{fig:modelpipeline}
  \end{minipage}

\end{figure}

We formulate tone style transfer as an in-context generation task and propose our ICTone model. As depicted in Fig.~\ref{fig:modelpipeline}, it takes both the content and reference images as input and implicitly learns the desired tonal transformation.
\subsubsection{In-Context Generation}
\mbox{}\\

Our goal is to learn a mapping function that transfers the tone characteristics from a reference image $I_r$ to the content image $I_c$ while preserving its semantic structure. Given a triplet $(I_c, I_r, I_t)$ where $I_t$ denotes the target image with the same tone as the reference image, the model is trained to approximate the conditional distribution $p_\theta(I_t\mid I_c, I_r)$. 

Unlike conventional tone transfer methods that explicitly disentangle content and style representations, recent progress in in-context generation~\cite{wu2025less,zhang2025context,li2025ic,li2026all} reveals that diffusion models are capable of understanding visual relationships directly from contextual examples. 
This observation motivates us to formulate tone transfer as a conditional inpainting problem.
Specifically, we construct a joint visual context by concatenating the content image and the reference style image $(I_c, I_r)$ along the spatial dimension. A binary mask is then applied to the region corresponding to the stylized output, while the original content and reference regions remain visible. The masked region is initialized as noise and serves as the target area~\cite{wen2025lucie}.
The diffusion transformer (DiT) is trained to predict and progressively denoise the masked region conditioned on the visible context. The model learns to preserve the structural information from $I_c$ while adapting the tonal characteristics observed in $I_r$. In this way, tone-consistent stylization emerges implicitly through contextual reasoning, without requiring explicit content-style disentanglement.
\subsubsection{Tone Reward Feedback Learning}
\label{sec:tone_reward}
\mbox{}\\
To further improve tonal fidelity, we introduce a tone reward feedback learning mechanism to constrain the generation process to align outputs with the reference tone distribution.
Specifically, we incorporate reward feedback learning inspired by the ReFL framework~\cite{xu2023imagereward}, which optimizes text-to-image synthesis toward reference-conditioned aesthetic alignment. Unlike ReFL, which optimizes text-to-image generation toward reference-conditioned aesthetic alignment, we extend this idea to the content–reference in-context generation setting.

For a triplet $s_i=(I_c^i, I_r^i, I_t^i)$, the generator produces a stylized image $\hat{I}^i=I_\theta(I_c^i, I_r^i)$. A pretrained tone style scorer $\mathcal{M}_{\mathrm{TS\text{-}Scorer}}$ then evaluates the alignment between the generated image $\hat{I}^i$ and the reference tone image $I_r^i$, providing a fine-grained supervision signal. The reward loss is defined as:
\begin{equation}
	\mathcal{L}_{\mathrm{tone}}
	= \mathbb{E}_{s_i\sim\boldsymbol{\mathcal{S}}}\big[\phi\big(\mathcal{M}_{\mathrm{TS\text{-}Scorer}}(I_r^i,\, I_\theta(I_c^i, I_r^i))\big)\big]\!,
\end{equation}
where $\mathbf{\boldsymbol{\mathcal{S}}}=\{\boldsymbol{s_i}\}_{i=1}^n$ denotes the set of triplets, $\phi$ maps scorer outputs to per-sample loss values, and $I_\theta(\cdot)$ is the diffusion model parameterized by $\theta$. This design enforces explicit tone consistency during generation, thereby enhancing both the stability and fidelity of tone style transfer.

\subsubsection{Training Objectives}
\mbox{}\\
We adopt the flow-matching objective as the primary optimization target~\cite{zhao2026extending}. Concretely, the model is trained to predict the target velocity field $\boldsymbol{v}_t$ from a noised latent $\boldsymbol{s}_t$, with the objective:
\begin{equation}
	\mathcal{L}_{\mathrm{FM}}=\mathbb{E}_{\boldsymbol{s}_0,t,\epsilon}\|\boldsymbol{v}_\theta-\boldsymbol{v}_t\|^2,
\end{equation}
where 
$v_t = \frac{d\alpha_t}{dt} \, \boldsymbol{s}_0 + \frac{d\sigma_t}{dt} \, \varepsilon$, and $\alpha_t = 1 - t,\; \sigma_t = t$ are continuous-time coefficients with $t \in [0,\, 1]$. $\boldsymbol{v}_\theta$ denotes the predicted velocity. 
This formulation is consistent with recent flow-based generative paradigms, enabling direct supervision of the continuous transformation dynamics between data and noise distributions.

After the model achieves stable convergence under $\mathcal{L}_{\mathrm{FM}}$, we introduce an additional tone reward feedback loss described in Section~\ref{sec:tone_reward} to further encourage tone alignment.
The final training objective integrates flow matching with tone reward feedback loss and is defined as:
\begin{equation}
	\mathcal{L}=\mathcal{L}_\mathrm{FM}+\lambda_\mathrm{tone}\mathcal{L}_\mathrm{tone},
\end{equation}
where $\lambda_{\text{tone}}=0$ before training step $S$ and $\lambda_{\text{tone}}=1$ afterward.

\section{Experiments}
\subsection{Experimental Setup}
\noindent\textbf{Datasets.} 
We conduct experiments on two datasets: TST2K and PST50~\cite{gong2025sa}. 
TST2K is our self-constructed dataset that covers a broader range of real-world scenes and tonal styles, providing a more rigorous benchmark for evaluating transfer quality. PST50 is used as a generalization test, where models are directly evaluated without additional training. It contains 50 paired triplets and is used to assess cross-dataset transfer performance.

\noindent\textbf{Comparison Methods.}
We compare our method with leading methods from photorealistic style transfer, color transfer, and diffusion-based style transfer, as well as the closed-source image editing model Nano Banana2.

\noindent\textbf{Evaluation Metrics.} We evaluate all methods using the following  metrics:
content preservation (CP), color difference ($\Delta E$), deep color difference (CD), and aesthetic quality (Aes). Among them, $\Delta E$ and CD are designed to measure the fidelity and accuracy of tone style transfer.
Content Preservation (CP) is measured using the Structural Similarity Index Measure (SSIM) computed on LDC-generated~\cite{soria2022ldc} edge maps, evaluating structural similarity between the generated output and the stylized ground truth.
Color Difference ($\Delta E$)~\cite{luo2001development} is a widely adopted international standard for measuring perceptual color differences between the stylized and output images.
Deep Color Difference (CD) is a neural network–based metric designed to better align color difference estimation with human perceptual judgments~\cite{chen2023learning}.
Aesthetic Quality (Aes) is measured by the mean aesthetic score predicted by an image aesthetic assessment model~\cite{sheng2023aesclip}, reflecting the overall visual appeal and tonal harmony of the results.

\noindent\textbf{Implementation Details.}
The tone style scorer is trained in two stages. 
For the weakly supervised contrastive learning stage, we train the backbone and the projection layer for 50 epochs with learning rates of $1\times10^{-4}$ and $3\times10^{-3}$, respectively, and set the temperature parameter in Eq.~\ref{equ:WSL} to $\tau=0.1$. 
For the subsequent preference learning stage, the projection layer is fine-tuned for 2 epochs using 20K human-ranked data, with the triplet margin in Eq.~\ref{equ:tri} set to $m=0.3$. 
To better match real-world conditions, our training incorporates degradations such as color desaturation, exposure reduction, blurring, and Gaussian noise.
We adopt FLUX.1 Fill as the backbone and fine-tune it on our constructed triplet dataset using the LoRA method~\cite{hu2022lora}. Training is performed on four NVIDIA A100 GPUs with the AdamW optimizer, where the learning rate is set to $1\times10^{-4}$ and the weight decay to $1\times10^{-3}$. The model is optimized for a total of 50{,}000 iterations.

\subsection{Experimental Results}
\subsubsection{Quantitative Evaluation}
\mbox{}\\

\begin{table}[t]
	\small
	\setlength{\tabcolsep}{3pt}
	\renewcommand{\arraystretch}{1.05}
	\centering
	\caption{Quantitative comparisons with state-of-the-art methods on the TST2K and PST50 benchmarks. 
    ICTone achieves superior performance in color difference (CD and $\Delta E$) and aesthetic quality (Aes), while maintaining competitive content preservation (CP). 
    Methods marked with $^*$ denote retraining on our triplet dataset, which further improves their performance.}
	\resizebox{\linewidth}{!}{%
            \begin{tabular}{lllllllll}
			\hline
			\multirow{2}{*}{Method}      & \multicolumn{4}{c}{TST2K}                                                     & \multicolumn{4}{c}{PST50}                                             \\ \cmidrule(lr){2-5} \cmidrule(lr){6-9}
            &  CP\(\uparrow\)         & \(\Delta E \downarrow\) & CD \(\downarrow\)     & Aes\(\uparrow\)    & CP \(\uparrow\)    & \(\Delta E \downarrow\) & CD \(\downarrow\)    & Aes\(\uparrow\)    \\ \hline
			WCT$^2$~\cite{yoo2019photorealistic}            & 0.3952                & 15.13   & 4.991  & 0.7136  & 0.6482            & 16.34   & 5.574 & 0.2381 \\
			PhotoNAS~\cite{an2020ultrafast}                 & 0.6721                & 19.57   & 6.183  & 0.6347  & 0.6757             & 32.87   & 8.566 & 0.2293 \\
			MKL~\cite{pitie2007linear}                   & 0.7511                  & 12.67   & 4.694  & 0.7363  & 0.7618            & 16.12   & 5.241 & 0.2742 \\
			PhotoWCT~\cite{li2018closed}                 & 0.5852                   & 21.35   & 7.021  & 0.6788  & 0.6477           & 20.83   & 6.487 & 0.2279 \\
			DeepPreset~\cite{ho2021deep}                           & \underline{0.7728}       & 8.205   & 4.108  & 0.7567  & 0.6566              & 16.90   & 5.679 & 0.1147 \\
			ModFlows~\cite{larchenko2025color}            & 0.6908           & 12.43   & 4.945  & 0.7500  & 0.7334           & 15.52   & 5.651 & 0.2774 \\
			IPST~\cite{liu2023universal}                  & 0.7364          & 9.648   & 4.326  & 0.7081  & 0.7275           & 14.08   & 5.484 & 0.2344 \\
			RLPixTuner~\cite{wu2024goal}                  & 0.6815           & 25.61   & 7.514 & 0.6354  & 0.6789          & 21.51   & 7.445 & 0.1946 \\
			SA-LUT~\cite{gong2025sa}                      & 0.7082           & 11.10    & 4.929  & 0.6507  & \underline{0.7697}           & \textbf{11.25}   & \underline{4.483} & 0.2048 \\
			CAP-VSTNet~\cite{wen2023cap}                  & 0.7013           & 11.87   & 4.665  & 0.7175  & 0.7384            & 16.21   & 5.350  & 0.2705 \\
			CAP-VSTNet$^*$~\cite{wen2023cap}              & 0.7665           & 7.359   & \underline{3.663} & 0.7801 & 0.7545            & 14.45   & 4.937 & 0.2779 \\
			Neural Preset~\cite{ke2023neural}             & 0.7480           & 10.51   & 4.713  & 0.7551   & 0.6502           & 19.53   & 6.965 & 0.1454 \\
			Neural Preset$^*$~\cite{ke2023neural}         & 0.7707          & \underline{7.886}   & 3.860   & 0.7808  & 0.6914     & 17.81   & 6.157 & 0.1714 \\
            CSGO~\cite{xing2024csgo} &0.3440 &19.00 &7.166 &0.7209 &0.3836 &26.18 &7.747 &\textbf{0.3607} \\
            
            Nano Banana2~\cite{google2026nanobanana2} &0.5968 &12.78 &4.947 &\underline{0.7885} &0.5901 &17.19 &5.988 &0.2793 \\
			ICTone                                         & \textbf{0.8644}   & \textbf{5.776}   & \textbf{2.634}  & \textbf{0.7904}  & \textbf{0.7902}  & \underline{12.81}   & \textbf{4.448} & \underline{0.2820}  \\ \hline
			GT                                           & 1.000               & 0.000       & 0.000      & 0.7978  & 1.000                    & 0.000       & 0.000     & 0.3305 \\
            \hline
		\end{tabular}
	}
	\label{tab:sota_cmp}
\end{table}

\begin{table*}[t]
	\renewcommand{\arraystretch}{1.05}
	\centering
	\setlength{\tabcolsep}{2.5pt}
	\caption{Comparison of average rankings across different methods.}
	\resizebox{\linewidth}{!}{%
		\begin{tabular}{r|ccccccc}
			\toprule
			Method & CAP-VSTNet\cite{wen2023cap} &Neural Preset\cite{ke2023neural} & SA-LUT\cite{gong2025sa} &MKL\cite{pitie2007linear} &ModFlows\cite{larchenko2025color} & IPST\cite{liu2023universal} & ICTone \\
			\midrule
			Average Ranking $\downarrow$ &4.35 & 2.70&4.05 &3.60 &6.00 &5.20 &\textbf{1.00}   \\
			\bottomrule
		\end{tabular}
	}
	\label{tab:average_ranking}
\end{table*}

We conduct a comprehensive quantitative comparison between our ICTone and several SOTA approaches on the TST2K and PST50 benchmarks, where PST50 serves as a direct generalization benchmark without dataset-specific adaptation. The results are presented in Tab.~\ref{tab:sota_cmp}.

On TST2K, our ICTone achieves the best performance across all evaluation metrics, demonstrating superior tone alignment with reference styles and content preservation. ICTone achieves the lowest color difference and deep color difference, demonstrating high fidelity in reproducing the target tone style. Note that ICTone achieves the highest aesthetic score 0.7904, closely approaching the ground truth 0.7978, which suggests that the improvements in tone accuracy do not compromise visual appeal. The low CP of Nano Banana2 is mainly caused by reference foreground leakage, which introduces undesired reference content into the output and degrades content preservation.

We further test on PST50 to evaluate the generalization capability and ICTone achieves the best overall performance. It obtains the highest CP of 0.7902 and the lowest CD of 4.448. Although SA-LUT achieves a lower $\Delta E$ on this dataset, its CD and aesthetic score are inferior to those of ICTone. This indicates that ICTone maintains stronger perceptual tone fidelity and overall visual quality under cross-dataset evaluation. The consistent gains across both datasets further demonstrate the robustness and generalization capability of ICTone on data outside the training distribution. Notably, the overall aesthetic scores on PST50 are generally lower, which can be attributed to its relatively darker tone distributions that tend to receive lower scores from the aesthetic assessment model. CSGO is an exception because it is fine-tuned from a style transfer model and tends to produce excessive saturation enhancement. Such unnatural color enhancement can be favored by the aesthetic assessment model, leading to an Aes score even higher than GT, but it does not indicate better retouching fidelity.

\subsubsection{Qualitative Evaluation}
\mbox{}\\
\begin{figure*}[tb]
	\centering
	\includegraphics[width=1.0\linewidth]{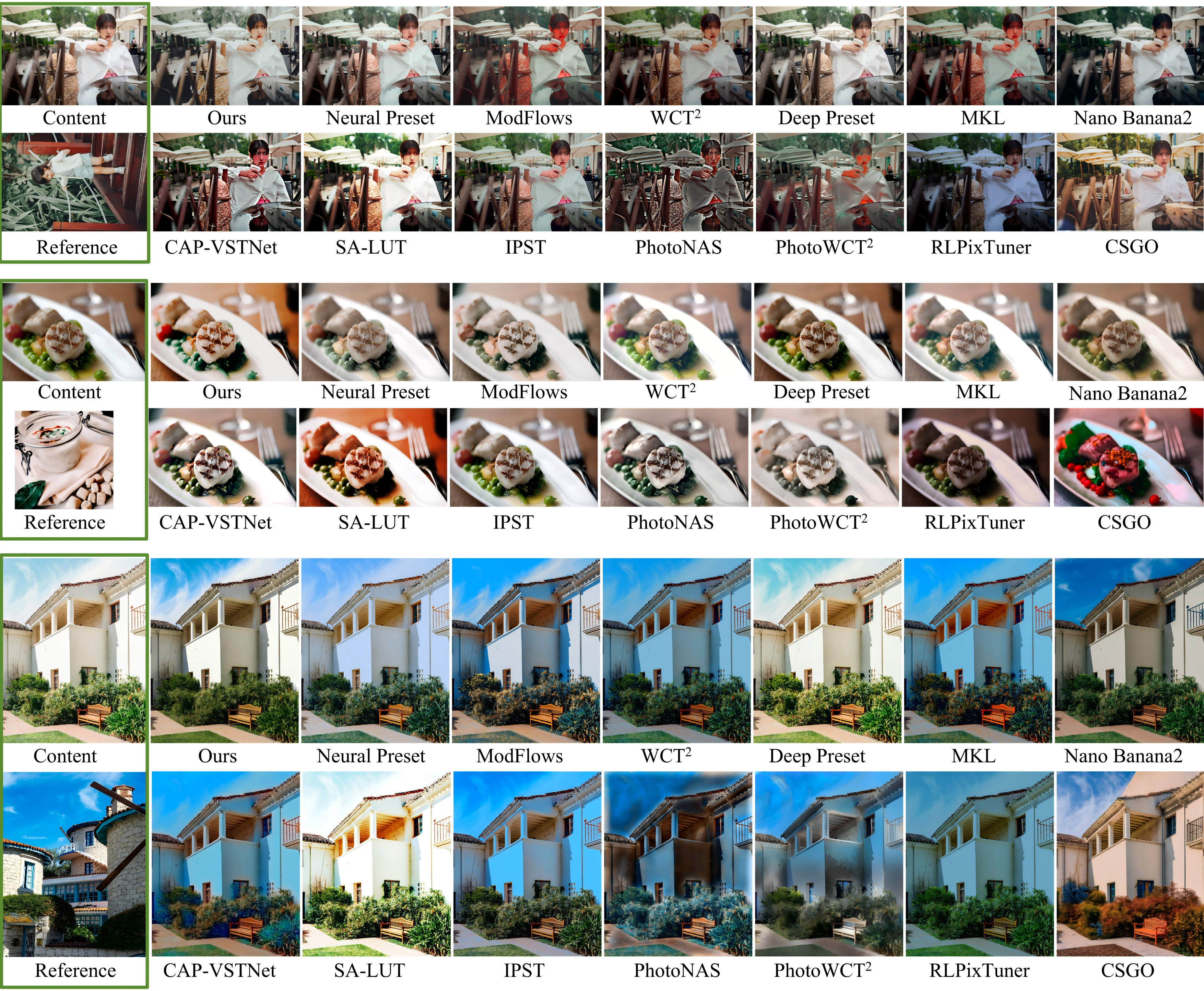}
	\caption{Qualitative visual comparisons of our method and the baseline methods.
	}
	\label{fig:vis_compare}
\end{figure*}

Fig.~\ref{fig:vis_compare} presents qualitative comparisons across portraits, urban scenes, and food images. ICTone consistently produces visually superior results with accurate tone style alignment and strong content awareness.
In portrait transfer (Fig.~\ref{fig:vis_compare}~(a)), prior methods often incorrectly transfer the warm hue of the railing onto the subject’s face, leading to noticeable color contamination. In contrast, our result faithfully reproduces the reference skin tone while preserving natural facial colors. In food image transfer (Fig.~\ref{fig:vis_compare}~(b)), our results closely match reference styles in color hierarchy and contrast, enhancing visual appeal. For urban scenes (Fig.~\ref{fig:vis_compare}~(c)), ICTone effectively captures both chromatic and illumination-level attributes, resulting in outputs with coherent lighting and tone gradation. More qualitative results are provided in the supplementary material.

\subsubsection{User Study}
\mbox{}\\
To assess the perceptual quality of stylization results, we conduct a user study following the protocols established in~\cite{gong2025sa,ke2023neural}. The study compares seven representative methods and uses 20 randomly sampled image pairs from the TST2K dataset. For each method, participants performed pairwise comparisons against the remaining six methods, resulting in 120 comparisons per method.

In each comparison, participants were presented with two stylized outputs and asked to select the one they perceived as superior in overall quality. Based on these selections, we compute the win rate for each method. Tab.~\ref{tab:average_ranking} summarizes the average ranking across 20 participants.

Our ICTone consistently achieved the top rank across all individual user rankings, reflecting the superior balance our approach achieves among structural fidelity, tone consistency and aesthetic coherence.

\subsection{Ablation Study}
\begin{table}[!tbp]
	\centering
	\small 
	\setlength{\tabcolsep}{4pt} 
	\caption{Ablation studies of high-quality dataset filtering.}
    \begin{tabular}{cccccc}
		\toprule
		Setup &  CP $\uparrow$ & $\Delta E \downarrow$ & CD $\downarrow$ &  AesScore $\uparrow$\\
		\midrule
		w/o filtering & 0.7950  &9.859  &3.674 & 0.7727 \\
		w/ filtering &\textbf{0.8567}  &\textbf{6.035}  &\textbf{2.737}  & \textbf{0.7834}  \\
		\bottomrule
	\end{tabular}
	\label{tab:ablation_data_rl}
\end{table}
\subsubsection{Effectiveness of Large-scale Triplet Dataset}
\mbox{}\\
To evaluate the effectiveness of our proposed dataset, we retrain two representative tone transfer models, CAP-VSTNet~\cite{wen2023cap} and Neural Preset~\cite{ke2023neural}, on TST100K. The retrained variants are denoted as CAP-VSTNet$^*$ and Neural Preset$^*$, respectively. Quantitative results are reported in Tab.~\ref{tab:sota_cmp}.

Both retrained models exhibit notable performance gains compared to their original counterparts, particularly in color difference and aesthetic quality. This demonstrates that large-scale triplet supervision is crucial for tone style transfer.

In addition, we validate the effectiveness of the high-quality filtering used in the construction of the TST100K dataset. As depicted in Table 3, the results demonstrate that filtering out inconsistent and aesthetically unpleasing reference–stylized pairs during large-scale data construction significantly enhances model performance. The experiment was conducted without incorporating reward learning. This finding highlights the importance of high-quality data curation for achieving robust tone style transfer.

\subsubsection{Effectiveness of the Tone Style Scorer}
\mbox{}\\
\begin{table}[t]
    \centering
    \caption{\textbf{Filter-based image retrieval evaluation on the IFFI dataset.} TS-WCL denotes our Tone Scorer trained with weakly supervised contrastive learning, and TS-WCL-PL further incorporates preference learning. 
    }
    \setlength{\tabcolsep}{2pt} 
    \small 
    \begin{tabular}{lcccc}
    \toprule
    Method        & Recall@1 $\uparrow$ & Recall@2 $\uparrow$ & Recall@5 $\uparrow$ & PAcc $\uparrow$ \\ \midrule
    VGG Gram~\cite{johnson2016perceptual}        &34.00                     &47.75                     &67.44                   &58.21         \\
    Neural Disc~\cite{ke2023neural}        &34.63                     &48.19                     &69.94                   &65.12         \\
    CSD~\cite{somepalli2024measuring}             &63.13                     &73.75                     &87.50                   &70.89       \\
    TS-WCL         &93.19                     &97.44                     &99.38                   &74.59        \\ 
    TS-WCL-PL      &\textbf{97.50}                     &\textbf{99.56}                     &\textbf{100.0}                   &\textbf{82.67}        \\ \bottomrule
    \end{tabular}
    \label{tab:DiscCompare}
\end{table}
\begin{table}[!tbp]
	\centering
	\setlength{\tabcolsep}{4pt} 
	\caption{Ablation studies of Reward Learning.
    WCL indicates the reward model trained with Weakly-supervised Contrastive Learning only, while PL indicates the reward model further fine-tuned with Preference Learning.}
	\small
    \begin{tabular}{cccccc}
        \toprule
        WCL & PL & CP $\uparrow$ & $\Delta E \downarrow$ & CD $\downarrow$ &  AesScore $\uparrow$\\
        \midrule
        - & - &0.8567  &6.035  &2.737  & 0.7834 \\
        \ding{51} & - & 0.8572  & 5.795& 2.765 &0.7871  \\
        \ding{51} & \ding{51} & \textbf{0.8644}  & \textbf{5.776} & \textbf{2.634} & \textbf{0.7904} \\
        \bottomrule
    \end{tabular}
	\label{tab:ablation_reward_learning}
\end{table}
To evaluate the effectiveness of our tone style scorer in perceiving preset styles, we conduct an image filter retrieval experiment on the IFFI dataset\cite{Kinli_2021_CVPR}, as shown in Tab.~\ref{tab:DiscCompare}. We compare the tone style scorer with prior style similarity methods, including VGG Gram~\cite{johnson2016perceptual}, Neural Disc~\cite{ke2023neural} and CSD~\cite{somepalli2024measuring}. TS-WCL is our tone style scorer trained using only tone style similarity learning, while TS-WCL-PL is the model further fine-tuned with 20K human-ranked pairs.

We evaluate the tone style scorer from two perspectives: retrieval accuracy and preference alignment. Preference accuracy (PAcc) measures alignment between model-predicted and human-preferred triplet rankings. CSD achieves stronger performance than VGG Gram and Neural Disc, but it still falls behind TS-WCL, especially in Recall@1 and PAcc, indicating that general style similarity is less effective than tone-specific representation learning for capturing fine-grained tonal differences. TS-WCL-PL further improves performance, reaching 100\% Recall@5 and 82.67\% PAcc. These results validate the effectiveness of our two-stage training strategy and confirm the effectiveness of the tone style scorer in filtering inconsistent reference–stylized pairs.

\subsubsection{Effectiveness of Tone Reward Learning}
\mbox{}\\
Tab.~\ref{tab:ablation_reward_learning} summarizes the fine-tuning results of ICTone with different tone style scorers. Comparing the first and second rows, we observe a clear improvement in $\Delta E$, indicating that tone similarity learning enables the scorer to better capture global style characteristics and effectively guide ICTone via reward learning. Further, comparing the second and third rows, all metrics show consistent gains, demonstrating that incorporating human preference signals allows the scorer to refine ICTone with improved style consistency and enhanced perceptual quality.
\section{Conclusion}
This paper presents TST100K and TST2K, the first large-scale datasets providing content-reference-stylized triplets for training and benchmarking tone style transfer. A two-stage tone style scorer is trained to ensure high quality and tone consistency during dataset construction. Leveraging the semantic priors of large generative models, we propose ICTone, a DiT framework that treats content and reference images in an in-context manner. Extensive experiments demonstrate that ICTone achieves perceptually consistent tone transfer when facing significant structural and semantic discrepancies between input images, achieving state-of-the-art in both quantitative metrics and human evaluations.

\noindent\textbf{Limitations and future work: }Our dataset construction pipeline is limited by the stylistic bias of existing aesthetic assessment models. Stronger aesthetic models could further improve the filtering of high-quality personalized styles. At the model level, our current strategy concatenates the content and reference images as input, which is not computationally efficient. More efficient conditioning mechanisms for reference-guided tone transfer are also worth exploring.

\section*{Acknowledgments}
\begin{sloppypar}
This research was supported by the Fund of the Shenzhen Science and Technology Program (JCYJ20250604184027034, QNXMB20250701090801002), Guangdong Basic and Applied Basic Research Foundation (2026A1515011435), the National Natural Science Foundation of China (Grant No. 62576177), the Fundamental Research Funds for the Central Universities 070-63263247, and the Tianjin Key Laboratory of Visual Computing and Intelligent Perception (VCIP). Computation is supported by the Supercomputing Center of Nankai University (NKSC). ``Science and Technology Yongjiang 2035'' key technology breakthrough plan project (2025Z053), Chinese government-guided local science and technology development fund projects (scientific and technological achievement transfer and transformation projects) (254Z0102G), National Science Fund of China under Grant No. 62361166670.
\end{sloppypar}
\bibliographystyle{splncs04}
\bibliography{main}
\newpage
\appendix
\renewcommand{\thesection}{\Alph{section}}
\renewcommand{\theHsection}{\Alph{section}}
\renewcommand{\theHsubsection}{\theHsection.\arabic{subsection}}
\renewcommand{\theHsubsubsection}{\theHsubsection.\arabic{subsubsection}}
\begin{center}
    \LARGE\bfseries Appendix
\end{center}
\addcontentsline{toc}{section}{Appendix}
\begin{itemize}
\setlength{\itemsep}{4pt}
\hrule
\vspace{3pt}
\item \textbf{\hyperref[sec:suppl_dataset_construction]{Dataset Construction}} \dotfill \pageref{sec:suppl_dataset_construction}
    \begin{itemize}
        \item \hyperref[sec:suppl_preset]{Preset-Based Data Generation} \dotfill \pageref{sec:suppl_preset}
        \item \hyperref[sec:suppl_tone]{Tone Style Scorer Filtering} \dotfill \pageref{sec:suppl_tone}
        \item \hyperref[sec:suppl_aesthetic]{Aesthetic Filtering} \dotfill \pageref{sec:suppl_aesthetic}
        \item \hyperref[sec:suppl_triplets]{Example Triplets from TST2K} \dotfill \pageref{sec:suppl_triplets}
    \end{itemize}

\item \textbf{\hyperref[sec:suppl_tone_training]{Tone Style Scorer Training Details}} \dotfill \pageref{sec:suppl_tone_training}
    \begin{itemize}
        \item \hyperref[sec:suppl_network]{Network Architecture} \dotfill \pageref{sec:suppl_network}
        \item \hyperref[sec:suppl_human]{Human Ranking Data} \dotfill \pageref{sec:suppl_human}
        \item \hyperref[sec:suppl_tone_train]{Training Details} \dotfill \pageref{sec:suppl_tone_train}
    \end{itemize}

\item \textbf{\hyperref[sec:suppl_implementation]{Implementation Details}} \dotfill \pageref{sec:suppl_implementation}
    \begin{itemize}
        \item \hyperref[sec:suppl_model_arch]{Model Architecture} \dotfill \pageref{sec:suppl_model_arch}
        \item \hyperref[sec:suppl_training_settings]{Training Settings} \dotfill \pageref{sec:suppl_training_settings}
        \item \hyperref[sec:suppl_infer]{Inference Details} \dotfill \pageref{sec:suppl_infer}
    \end{itemize}

\item \textbf{\hyperref[sec:suppl_addition_exp]{Additional Experimental Results}} \dotfill \pageref{sec:suppl_addition_exp}
    \begin{itemize}
        \item \hyperref[sec:suppl_details_metrics]{Details of Evaluation Metrics}  \dotfill \pageref{sec:suppl_details_metrics}
        \item \hyperref[sec:suppl_eval_metrics]{Additional Evaluation Metrics} \dotfill \pageref{sec:suppl_eval_metrics}
        \item \hyperref[sec:suppl_quant]{Quantitative Results} \dotfill \pageref{sec:suppl_quant}
        \item \hyperref[sec:suppl_user]{Details about User Study} \dotfill \pageref{sec:suppl_user}
        \item \hyperref[sec:suppl_visual]{Visual Results} \dotfill \pageref{sec:suppl_visual}
        \item \hyperref[sec:suppl_colorization_task]{Extension to Colorization Tasks} \dotfill \pageref{sec:suppl_colorization_task}
    \end{itemize}
\item \textbf{\hyperref[sec:suppl_limit]{Discussion and Limitations}} \dotfill \pageref{sec:suppl_limit}
\vspace{5pt}
\hrule
\end{itemize}
\begin{figure}[!tbh]
	\centering
	\includegraphics[width=1.0\linewidth]{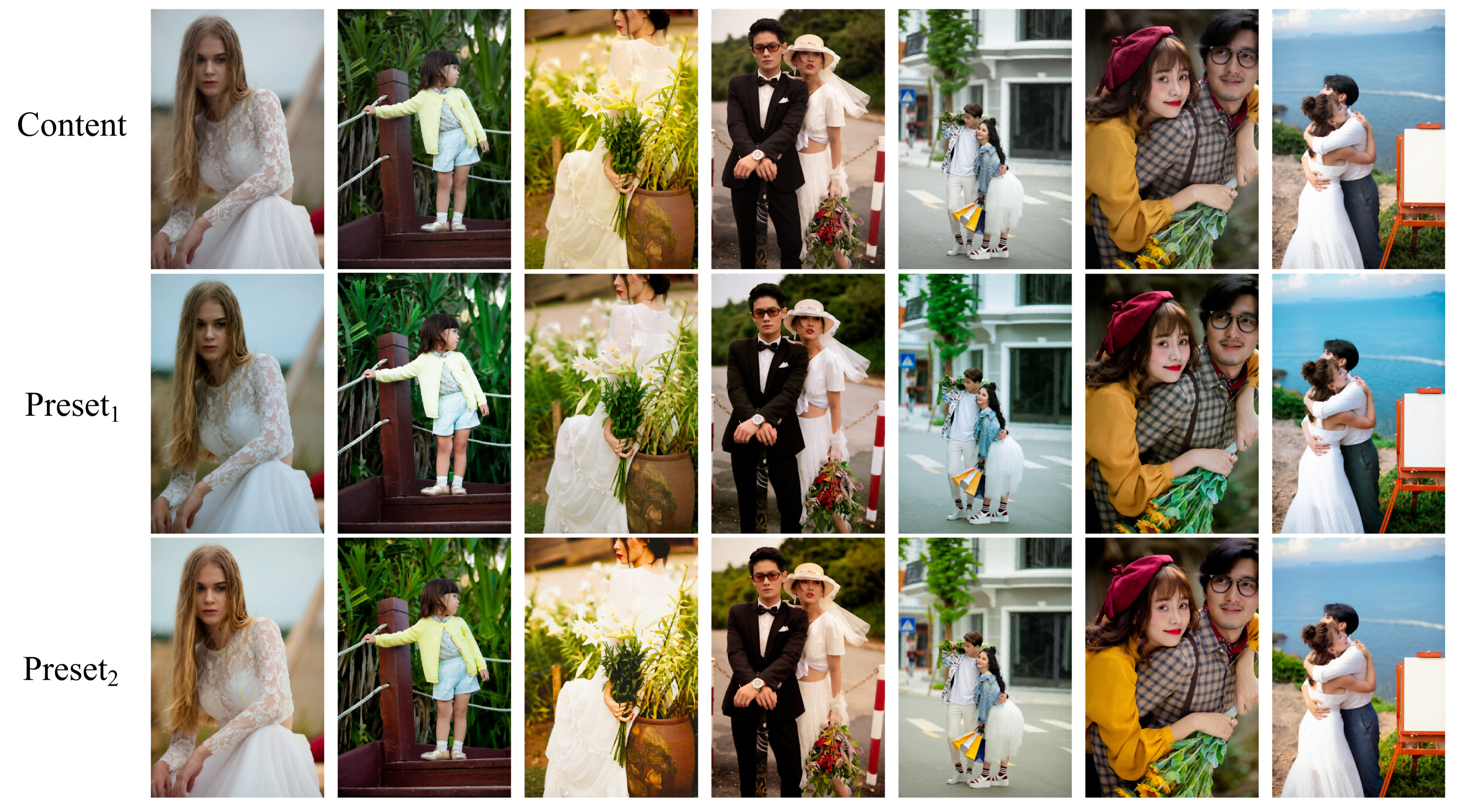}
	\caption{Examples of preset-generated candidates. The same preset applied to different content images can produce stylized results with similar tone style, but it may also yield inconsistent tonal effects, resulting in noisy candidate pairs.}
	\label{fig:showcase_preset}
    \vspace{-6mm}
\end{figure}

\section{Dataset Construction}
\label{sec:suppl_dataset_construction}
\subsection{Preset-Based Data Generation}
\label{sec:suppl_preset}
Preset-based generation provides a scalable way to construct stylized image candidates. 
Given a content image and a preset, a stylized image can be obtained by directly applying the preset parameters. 
However, the tonal effect of a preset is highly dependent on the content image. 
When the same preset is applied to different images, it may produce stylized results with similar tone style, but it can also lead to inconsistent tonal effects. 
As illustrated in Fig.~\ref{fig:showcase_preset}, some preset-generated results exhibit consistent tonal characteristics, while others deviate noticeably from the expected style. 
Moreover, different presets applied to different content images may still produce visually similar tone styles. 
For example, the image in the third column of the second row and the image in the second column of the third row exhibit highly similar tonal characteristics even though they are generated with different presets. 
This further increases the ambiguity of preset-generated candidates. 
This phenomenon introduces noise into preset-generated candidates and makes it difficult to ensure reliable tone alignment. 
Therefore, an additional mechanism is required to identify candidate pairs with consistent tone style.
\begin{figure}[!th]
	\centering
	\includegraphics[width=1.0\linewidth]{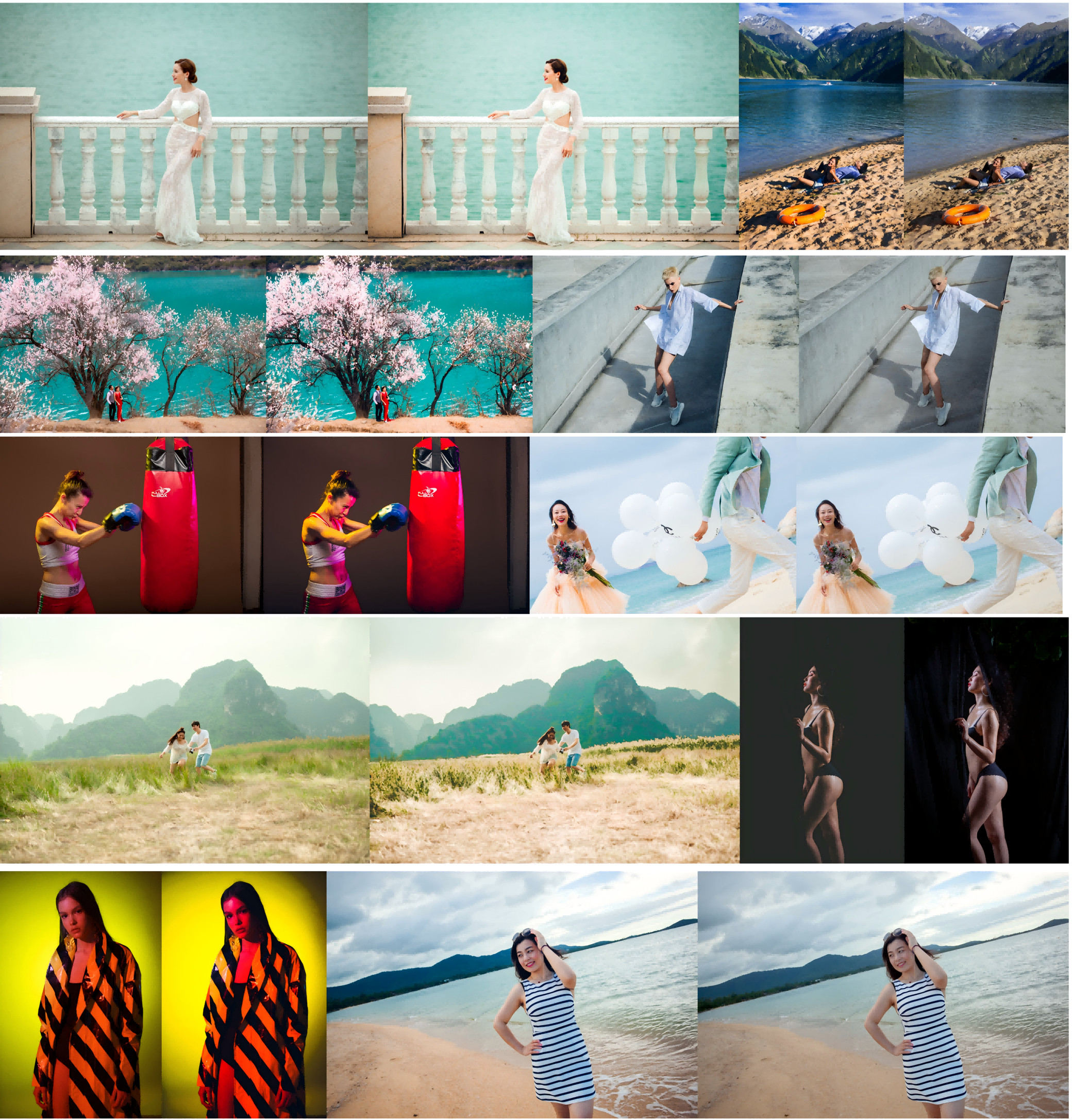}
	\caption{Examples of preset-generated pairs with inconsistent tone styles identified by the tone style scorer. Despite using the same preset, the stylized results exhibit different perceptual tone styles. In each example, the first and last images are the original images, and the two middle images are the stylized results after applying the preset.}
    \vspace{-6mm}
	\label{fig:showcase_tone_filtering}
\end{figure}
\subsection{Tone Style Scorer Filtering}
\label{sec:suppl_tone}
Preset-based generation produces a large number of stylized candidates, but many of them exhibit inconsistent tone styles even when generated using the same preset. 
This inconsistency mainly arises from the interaction between preset parameters and image content. 
As a result, preset-generated pairs often contain noisy samples that do not share consistent perceptual tone styles.

To address this issue, we employ the tone style scorer to evaluate the tone similarity between stylized images and their corresponding reference images. 
The scorer filters out samples with low tone similarity scores and retains only pairs with consistent perceptual tone styles.

Fig.~\ref{fig:showcase_tone_filtering} shows examples of filtered pairs identified by the tone style scorer. 
In each example, the same preset is applied to two different content images. 
Although the preset parameters are identical, the resulting stylized images exhibit noticeably different perceptual tone styles. 
For each example, the first and last images show the original content images before applying the preset, while the two middle images show the corresponding stylized results. 
These examples illustrate that preset-based generation alone cannot guarantee consistent tone styles and highlight the necessity of tone style scorer filtering.

\subsection{Aesthetic Filtering}
\label{sec:suppl_aesthetic}
Preset-based stylization may occasionally degrade the visual quality of the image. 
Although presets are designed to enhance tonal appearance, some combinations of presets and image content produce results with lower aesthetic quality than the original image.

To prevent such cases, we introduce an aesthetic assessment model to filter low-quality samples. 
For each stylized candidate, the aesthetic score is compared with that of the original content image. 
Candidates with aesthetic scores lower than the original image are removed.

\begin{figure}[!h]
	\centering
	\includegraphics[width=0.8\linewidth]{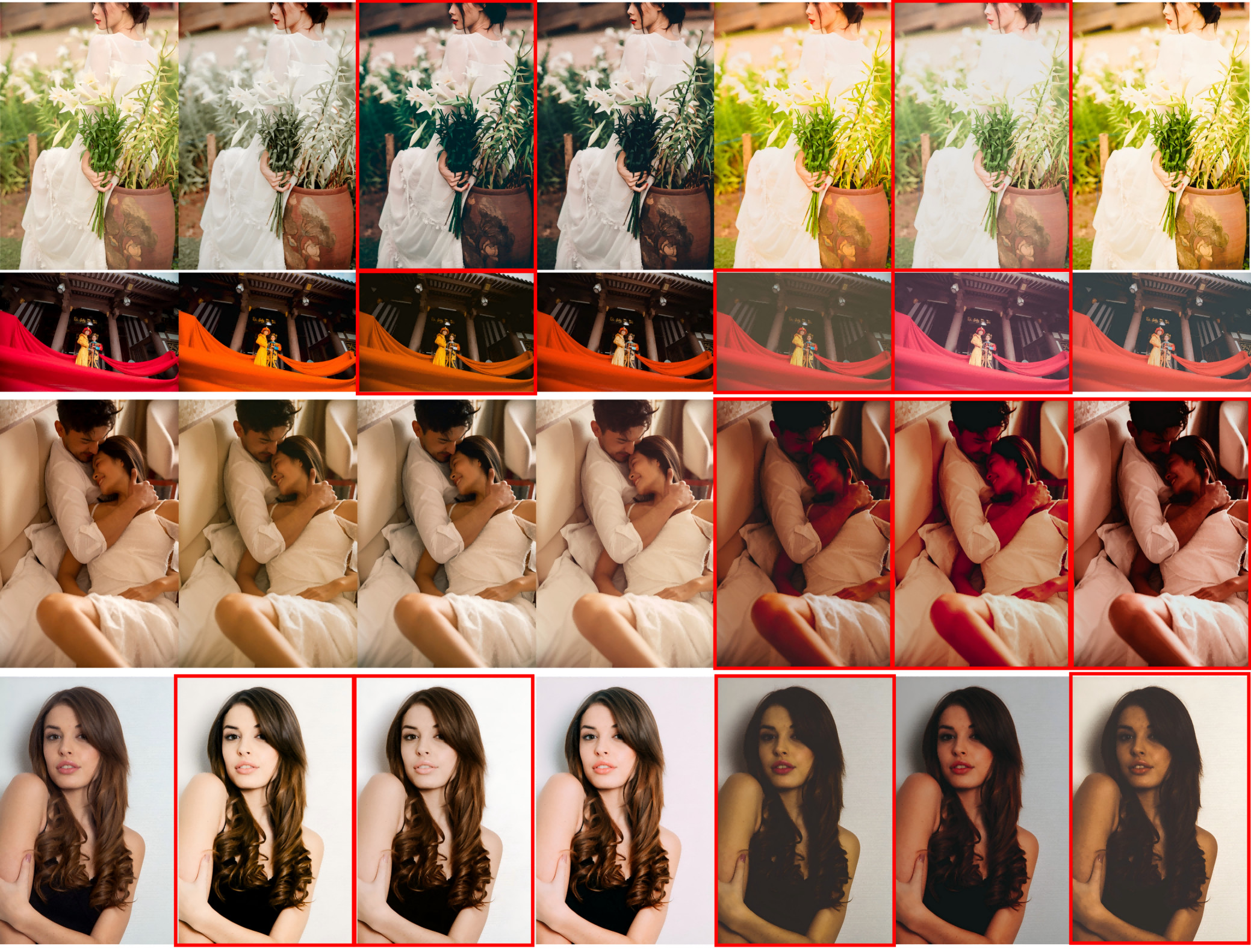}
	\caption{Examples of aesthetic filtering. The first image in each row is the content image, followed by stylized results generated with different presets. Images highlighted with red boxes have lower aesthetic scores than the original image and are removed by the aesthetic assessment model.}
    \vspace{-6mm}
	\label{fig:showcase_aes_filtering}
\end{figure}

Examples are shown in Fig.~\ref{fig:showcase_aes_filtering}. 
In each row, the first image is the content image and the remaining images are stylized results obtained using different presets. 
Some stylized results reduce the overall visual quality, such as producing unnatural colors or poor tonal balance. 
These cases are identified by the aesthetic assessment model and removed from the dataset.

\subsection{Example Triplets from TST2K}
\label{sec:suppl_triplets}
\begin{figure}[!t]
	\centering
	\includegraphics[width=1.0\linewidth]{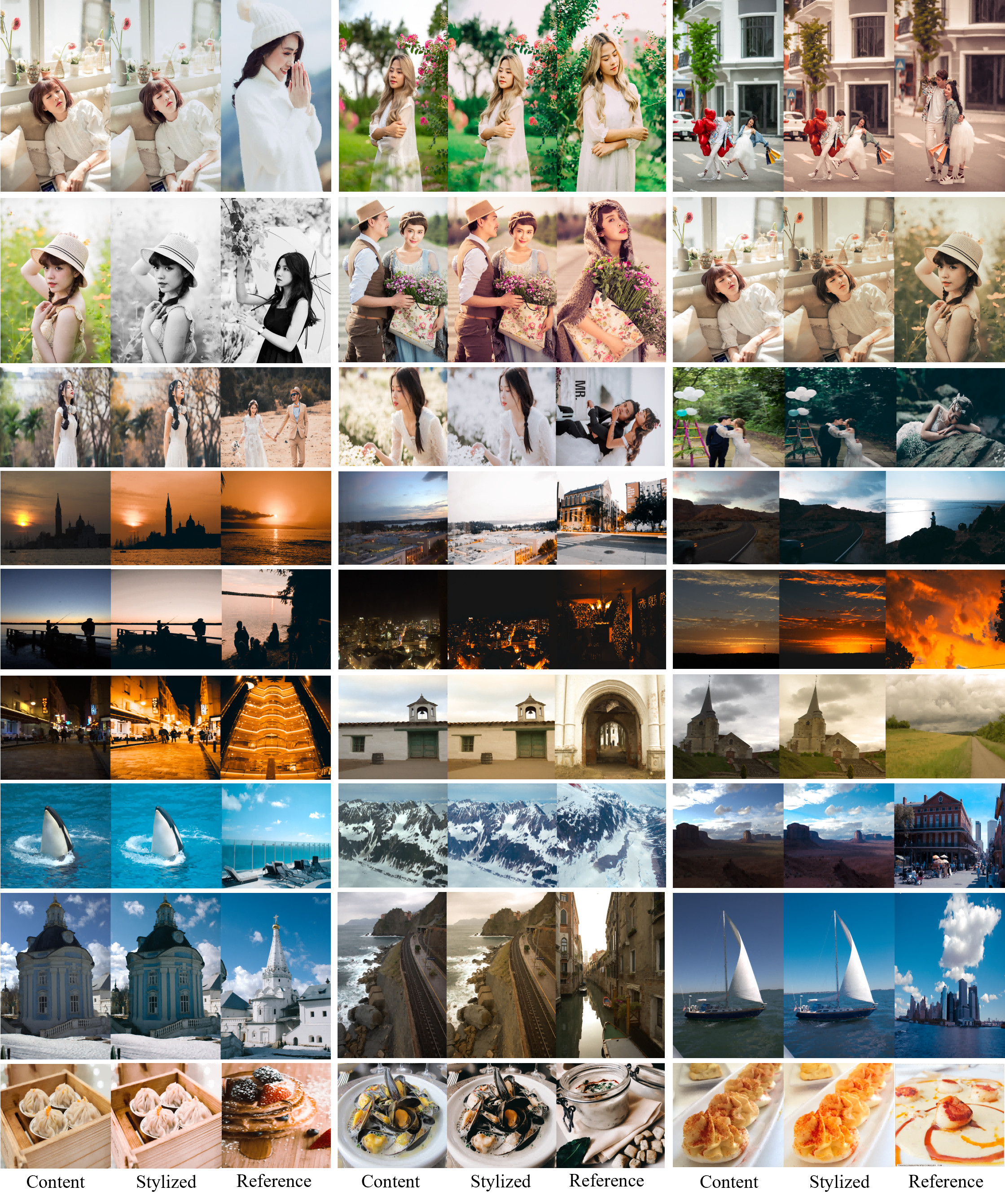}
	\caption{Example triplets from our TST2K benchmark. 
		Each triplet consists of a content image, a stylized image generated from the content, and a reference image providing the target style. This arrangement facilitates direct comparison between the original content, 
		the stylized output, and the style image.
		}
	\label{fig:showcase_triplet}
\end{figure}
To better illustrate the construction of our dataset, Fig.~\ref{fig:showcase_triplet} shows additional examples of triplets from TST2K. Each triplet consists of a content image, a reference image, and the corresponding stylized output.
This triplet design explicitly models the relationship between semantic preservation and style transfer. By jointly considering content fidelity, stylistic guidance, and the resulting stylized output, our dataset enables more robust training and fair evaluation of models that aim to preserve semantics while achieving effective tone style transfer.
\section{Tone Style Scorer Training Detail}
\label{sec:suppl_tone_training}
\subsection{Network Architecture}
\label{sec:suppl_network}

The tone style scorer is built upon a CLIP-initialized Vision Transformer following the general design of Contrastive Style Descriptors~\cite{somepalli2024measuring}. Specifically, each input image is first encoded by a ViT-B/16 visual backbone, and the resulting global image representation is further mapped into a tone-aware embedding space by a projection head.

Given a reference image $I_r$ and a stylized image $I_s$, the two images are processed by a shared-weight ViT backbone. Let $h_r$ and $h_s$ denote the global features extracted by the ViT-B/16 backbone. A two-layer MLP projection head is then applied to obtain the final 512-dimensional embeddings:
\begin{equation}
z_r = g(h_r), \qquad z_s = g(h_s),
\end{equation}
where $g(\cdot)$ denotes the projection head. The output embeddings are $\ell_2$-normalized before similarity computation. The final tone style similarity score is defined by cosine similarity:
\begin{equation}
s(I_r, I_s) = \frac{z_r^\top z_s}{\|z_r\| \|z_s\|}.
\end{equation}
\subsection{Human Ranking Data}
\label{sec:suppl_human}
To obtain high-quality supervision for preference learning, we construct a human-ranked dataset that focuses on ambiguous tone similarity cases. We first identify candidate samples where the similarity rankings predicted by different backbone-based tone scorers are inconsistent. The scorers are built on three representative architectures, including VGG, ResNet, and Vision Transformer. These disagreement cases often correspond to subtle tonal differences that are difficult for models to evaluate reliably.

For each annotation instance, we form a ranking group consisting of one anchor image and four candidate images. Human annotators are asked to rank the four candidates according to their tone similarity to the anchor image, from the most similar to the least similar.
By concentrating human annotation on samples where model predictions disagree, the resulting ranking data provides informative preference signals that help the tone style scorer better align with human perception.
\subsection{Training Details}
\label{sec:suppl_tone_train}
The tone style scorer is trained in two stages: weakly-supervised contrastive learning using preset-generated data and preference learning with human ranking annotations. In the first stage, the backbone and the projection layer are jointly trained for 50 epochs with a learning rate of $1\times10^{-4}$ for the backbone and $3\times10^{-3}$ for the projection layer. The batch size is set to 512. In the second stage, the scorer is further aligned with human perception using 20K human-ranked samples. Following the training strategy of ImageReward~\cite{xu2023imagereward}, each ranking group is converted into pairwise preference comparisons, and only the projection layer is fine-tuned for 2 epochs with a batch size of 128 while keeping the backbone fixed. To better match real-world conditions, we apply several appearance degradations during training, including color desaturation, exposure reduction, blurring, and Gaussian noise. The margin and temperature in the contrastive loss are set to $m=0.3$ and $\tau=0.1$, respectively.
\section{Implementation Details}
\label{sec:suppl_implementation}
\subsection{Model Architecture}
\label{sec:suppl_model_arch}
Our transfer model is built upon FLUX.1 Fill, a 12B rectified flow transformer originally designed for image completion and outpainting. FLUX models are based on a hybrid architecture of multimodal and parallel diffusion transformer blocks, which consists of a variational autoencoder and an MMDiT. The VAE first maps the input image into a compact latent representation, and the transformer then performs iterative denoising in latent space to predict the target visual content.

To adapt FLUX.1 Fill for tone style transfer, we reformulate the task as a masked outpainting problem. Specifically, we construct a joint visual context by concatenating the content image and the reference style image, $(I_c, I_r)$, along the spatial dimension. A binary mask is then applied to the target region for stylized generation, while the original content and reference regions remain visible as contextual guidance. The masked region is initialized with noise and serves as the generation target. In this way, the model can jointly access the source semantics from the content image and the desired tone style from the reference image within a unified visual canvas.

To reduce memory consumption and improve inference efficiency, we remove the original text-conditioning branch of FLUX.1 Fill and retain only the latent image pathway together with the transformer denoiser. In practice, the concatenated visual context is encoded by the VAE into latent tokens, which are then processed by the Flux transformer backbone for masked latent prediction. The predicted latent output is finally decoded by the VAE decoder to obtain the stylized image. This design preserves the strong image editing prior of FLUX.1 Fill while making the model more suitable for purely visual tone transfer.
\subsection{Training Settings}
\label{sec:suppl_training_settings}
The transfer model is initialized from FLUX.1 Fill and fine-tuned on the constructed triplet dataset using LoRA~\cite{hu2022lora}. For each training sample, the content image and the reference style image are concatenated horizontally to form a single-row input canvas. The region corresponding to the stylized output is masked and used as the prediction target. During fine-tuning, only the diffusion transformer layers are updated. LoRA is applied with a rank of 256 and an alpha of 512. Training is conducted on four NVIDIA A100 GPUs using the AdamW optimizer, with a batch size of 4, a learning rate of $1\times10^{-4}$, and a weight decay of $1\times10^{-3}$. The model is optimized for a total of 50{,}000 iterations.
\subsection{Inference Details}
\label{sec:suppl_infer}
During inference, the input layout is kept the same as in training. Specifically, the content image and the reference style image are horizontally concatenated to form a single-row visual canvas, and the target stylized region is masked for generation. To accelerate inference, we perform sampling with only 4 denoising steps, while setting the guidance scale to 50 following ICEdit~\cite{zhang2025context}.

Diffusion-based generative models are typically limited to moderate image resolutions. Directly applying ICTone to ultra-high-resolution images such as 4K or 8K is therefore computationally expensive. To address this limitation, a LUT-based post-processing strategy is adopted to extend the learned tone transformation to high-resolution images.

Given a content image $I_c$ and a reference image, ICTone first produces a stylized result at a resolution of 512. A 3D lookup table is then estimated from the content image and the generated stylized image to approximate the global color transformation learned by the model. The lookup table uses a $33 \times 33 \times 33$ lattice defined in the RGB color space. Each lattice vertex stores a transformed RGB value, and colors between lattice vertices are computed using trilinear interpolation.

To estimate the lookup table, dense color correspondences are collected from the paired pixels of the content image and the stylized image. Each pixel provides an input color vector $\mathbf{c}$ from the content image and a target color vector $\mathbf{s}$ from the stylized image. These pairs describe the color transformation induced by ICTone. The lookup table parameters are optimized by minimizing the reconstruction error between the transformed colors and the stylized colors
\begin{equation}
\min_{\mathrm{LUT}} \sum_i \| \mathrm{LUT}(\mathbf{c}_i) - \mathbf{s}_i \|_2^2 .
\end{equation}
\section{Additional Experiments}
\label{sec:suppl_addition_exp}
\subsection{Details of Evaluation Metrics}
\label{sec:suppl_details_metrics}
We evaluate all methods using four metrics, including content preservation (CP), color difference ($\Delta E$), deep color difference (CD), and aesthetic quality (Aes). Among them, $\Delta E$ and CD are used to evaluate the fidelity of tone style transfer, CP measures structural consistency, and Aes reflects the overall visual appeal of the generated results.

\paragraph{Content Preservation.}
Content preservation is measured by the Structural Similarity Index Measure (SSIM) computed on edge maps generated by LDC~\cite{soria2022ldc}. Let $I_o$ denote the output image and $I_{gt}$ denote the stylized ground truth. We first extract their edge maps using the LDC edge detector $\mathcal{E}(\cdot)$, and then compute SSIM between the two edge representations:
\begin{equation}
\mathrm{CP} = \mathrm{SSIM}\big(\mathcal{E}(I_o), \mathcal{E}(I_{gt})\big).
\end{equation}
Here, SSIM is defined as
\begin{equation}
\mathrm{SSIM}(x,y) =
\frac{(2\mu_x\mu_y + c_1)(2\sigma_{xy} + c_2)}
{(\mu_x^2 + \mu_y^2 + c_1)(\sigma_x^2 + \sigma_y^2 + c_2)},
\end{equation}
where $\mu_x$ and $\mu_y$ are the mean intensities, $\sigma_x^2$ and $\sigma_y^2$ are the variances, $\sigma_{xy}$ is the covariance, and $c_1,c_2$ are small constants for numerical stability. By computing SSIM on edge maps rather than RGB pixels, CP focuses on structural layout and contour consistency while reducing the influence of color changes introduced by style transfer.

\paragraph{Color Difference.}
We use the CIEDE2000 color difference~\cite{luo2001development}, which is an international standard for perceptual color difference measurement. Given corresponding pixels in the output image $I_o$ and the stylized ground truth $I_{gt}$, their colors are first converted from RGB to the CIELAB space, producing $(L_1^*,a_1^*,b_1^*)$ and $(L_2^*,a_2^*,b_2^*)$. The color difference at each pixel is then computed as

{\small
\begin{equation}
\Delta E_{2000} =
\sqrt{
\left(\frac{\Delta L'}{k_L S_L}\right)^2 +
\left(\frac{\Delta C'}{k_C S_C}\right)^2 +
\left(\frac{\Delta H'}{k_H S_H}\right)^2 +
R_T
\left(\frac{\Delta C'}{k_C S_C}\right)
\left(\frac{\Delta H'}{k_H S_H}\right)
},
\end{equation}
}

Here $\Delta L'$, $\Delta C'$, and $\Delta H'$ denote the corrected lightness difference, chroma difference, and hue difference. $S_L$, $S_C$, and $S_H$ are the corresponding weighting functions. $R_T$ is a rotation term that improves the modeling of hue differences in the blue region. $k_L$, $k_C$, and $k_H$ are parametric factors and are set to $1$ under standard conditions. 

The image-level color difference is computed by averaging the pixel-wise values over all spatial locations

{
\begin{equation}
\Delta E = \frac{1}{N}\sum_{p=1}^{N} \Delta E_{2000}^{(p)},
\end{equation}
}

where $N$ is the total number of pixels. A lower $\Delta E$ indicates that the generated result is closer to the stylized ground truth in perceptual color appearance.

\paragraph{Deep Color Difference.}
To better reflect human perception on photographic images, we further adopt the deep color difference metric proposed in~\cite{chen2023learning}. Unlike traditional handcrafted color difference formulae, this metric maps images into a learned feature space that is calibrated with human perceptual judgments. Let $\Phi(\cdot)$ denote the learned feature transform of the deep color difference model. The deep color difference between the output image and the stylized ground truth is computed as
\begin{equation}
\mathrm{CD} = \left\| \Phi(I_o) - \Phi(I_{gt}) \right\|_2.
\end{equation}
In practice, the metric is computed using the pretrained model released by~\cite{chen2023learning}. Compared with $\Delta E$, CD is more suitable for photographic images with complex semantics and local structures, and thus provides a complementary evaluation of tone transfer accuracy. Lower CD values indicate better alignment with the target tone style.

\paragraph{Aesthetic Quality.}
Aesthetic quality is measured by the mean predicted score from an image aesthetic assessment model~\cite{sheng2023aesclip}. Let $\mathcal{A}(\cdot)$ denote the aesthetic scorer. For a set of generated images $\{I_o^{(i)}\}_{i=1}^{M}$, the overall aesthetic quality is computed as
\begin{equation}
\mathrm{Aes} = \frac{1}{M}\sum_{i=1}^{M} \mathcal{A}\big(I_o^{(i)}\big),
\end{equation}
where $M$ is the number of test images. A higher Aes score indicates better overall visual appeal and more harmonious tonal presentation.
\subsection{Additional Evaluation Metrics}
\label{sec:suppl_eval_metrics}

\begin{table*}[t]
	\centering
	\small
	\caption{Quantitative comparison of content preservation and style similarity metrics on TST2K and PST50. CP$_{cnt}$ denotes content preservation computed between stylized and content images, CP denotes GT-based content preservation, TC$_{ref}$ denotes style similarity with respect to the reference image, and TC denotes GT-based style similarity. Our ICTone achieves the best overall performance, approaching GT upper bounds.}
	\begin{tabular}{lllllllll}
		\hline
		\multirow{2}{*}{Method}      & \multicolumn{4}{c}{TST2K}                         & \multicolumn{4}{c}{PST50}                  \\ \cmidrule(lr){2-5} \cmidrule(lr){6-9}
		& CP$_{cnt}$\(\uparrow\)      & CP\(\uparrow\)           & TC$_{ref}$\(\uparrow\)               & TC\(\uparrow\) & CP$_{cnt}$\(\uparrow\)     & CP\(\uparrow\)           & TC$_{ref}$\(\uparrow\)      & TC\(\uparrow\)  \\ \hline
		WCT$^2$~\cite{yoo2019photorealistic}           & 0.4116 & 0.3952          & 0.5655          & 0.5651 & 0.6818 & 0.6482          & 0.6797 & 0.6584 \\
		PhotoNAS~\cite{an2020ultrafast}             & 0.7193 & 0.6721          & 0.5429          & 0.5232 & 0.6901 & 0.6757          & 0.5375 & 0.5090  \\
		MKL~\cite{pitie2007linear}                 & 0.7809 & 0.7511          & 0.6353          & 0.6143 & 0.7305 & \underline{0.7618}          & 0.7646 & 0.7026 \\
		PhotoWCT~\cite{li2018closed}            & 0.6126 & 0.5852          & \underline{0.7030}           & 0.6700   & 0.6596 & 0.6477          & \underline{0.8156} & 0.7459 \\
		DeepPreset~\cite{ho2021deep}            & \textbf{0.8889} & \underline{0.7728}          & 0.6118          & 0.6635 & 0.7248 & 0.6566          & 0.5700   & 0.6197 \\
		ModFlows~\cite{larchenko2025color}            & 0.6949 & 0.6908          & 0.6876          & 0.6659 & 0.7740  & 0.7334          & 0.6929 & 0.6689 \\
		IPST~\cite{liu2023universal}                & 0.7749 & 0.7364          & 0.6231          & 0.6282 & 0.7576 & 0.7275          & 0.6827 & 0.6547 \\
		RLPixTuner~\cite{wu2024goal}          & 0.7324 & 0.6815          & 0.2497          & 0.2787 & \textbf{0.8081} & 0.6789          & 0.4775 & 0.5021 \\
		SA-LUT~\cite{gong2025sa}               & 0.6695 & 0.7082          & 0.5254          & 0.5573 & 0.7037 & 0.7697          & 0.7407 & \underline{0.7625} \\
		CAP-VSTNet~\cite{wen2023cap}          & 0.7148 & 0.7013          & 0.6245          & 0.6084 & 0.7040  & 0.7384          & 0.7693 & 0.7132 \\
		CAP-VSTNet$^*$~\cite{wen2023cap}    & 0.8303 & 0.7665          & 0.6505          & \underline{0.6717} & 0.7801 & 0.7545          & 0.7074 & 0.7312 \\
		Neural Preset~\cite{ke2023neural}       & 0.8314 & 0.7480           & 0.5249          & 0.5282 & 0.7420  & 0.6502          & 0.6086 & 0.5716 \\
		Neural Preset$^*$~\cite{ke2023neural} & \underline{0.8579} & 0.7707          & 0.6058          & 0.6248 & \underline{0.7919} & 0.6914          & 0.6087 & 0.6199 \\
		ICTone                & 0.7285 & \textbf{0.8644} & \textbf{0.8788} & \textbf{0.9280}  & 0.7323 & \textbf{0.7902} & \textbf{0.8799} & \textbf{0.8529} \\ \hline
		GT                           & 0.7472 & 1               & 0.8905          & 1      & 0.7209 & 1               & 0.8536 & 1      \\ \hline
	\end{tabular}
	\label{tab:supp_metrics}
\end{table*}

In the main paper, we report two key evaluation metrics: content preservation and style similarity, both computed between the stylized image and the ground-truth (GT) image. This GT-based protocol ensures that the metrics faithfully reflect how well the stylized output approximates the intended target.

For completeness, and following the practice of \emph{neural preset} evaluation, we also provide alternative metrics: CP$_{cnt}$, computed between the stylized image and the original content image, and TC$_{ref}$ computed between the stylized image and the reference image. As shown in Tab.~\ref{tab:supp_metrics}, the content preservation score between the GT and the content image is only 0.7472. This relatively low value arises from tone-style variations, which affect edge map extraction and consequently reduce metric accuracy. In contrast, GT-based evaluation provides a more reliable measure, as the GT naturally reflects the intended balance between content preservation and style transfer. While the alternative metrics offer complementary insights, the GT-based protocol appears to be a more consistent and informative choice for assessing the quality of stylized outputs.

From the quantitative comparison, we observe that our proposed \textbf{ICTone} achieves the highest style similarity (TC$_{ref}$) scores among all competing methods. Specifically, ICTone obtains style similarity scores close to the GT upper bound. These results demonstrate that ICTone consistently outperforms prior approaches, narrowing the gap to GT and validating its effectiveness in balancing semantic fidelity with stylistic accuracy.
\subsection{Additional Quantitative Results}
\label{sec:suppl_quant}
Following the SA-LUT evaluation protocol\cite{gong2025sa}, we further evaluate performance on TST2K using PSNR, SSIM~\cite{wang2004image}, and LPIPS~\cite{zhang2018unreasonable}. As shown in Tab.~\ref{table:psnr_ssim_lpips}, ICTone achieves the best performance across all three metrics, with the highest PSNR (25.83), the highest SSIM (0.94), and the lowest LPIPS (0.06).

\begin{table}[t]
	\centering
	\caption{Quantitative comparison on the TST2K dataset following the SA-LUT evaluation protocol. We report PSNR, SSIM, and LPIPS scores. Higher PSNR/SSIM and lower LPIPS indicate better perceptual quality.}
	\begin{tabular}{llll}
		\hline 
		Method& PSNR  & SSIM          & LPIPS         \\ \hline
		WCT$^2$~\cite{yoo2019photorealistic}  & 19.16 & 0.83          & 0.15          \\
		PhotoNAS~\cite{an2020ultrafast}       & 15.47 & 0.77          & 0.22          \\
		MKL~\cite{pitie2007linear}            & 20.19 & 0.85          & 0.13          \\
		PhotoWCT~\cite{li2018closed}          & 14.14 & 0.73          & 0.30          \\
		DeepPreset~\cite{ho2021deep}           & 22.85 & 0.89          & 0.14          \\
		ModFlows~\cite{larchenko2025color}     & 19.44 & 0.83          & 0.18          \\
		IPST~\cite{liu2023universal}        & 21.51 & 0.87          & 0.15          \\
		RLPixTuner~\cite{wu2024goal}          & 14.35 & 0.74          & 0.22          \\
		SA-LUT~\cite{gong2025sa}                 & 19.09 & 0.82          & 0.16          \\
		CAP-VSTNet~\cite{wen2023cap}          & 20.14 & 0.84          & 0.17          \\
		CAP-VSTNet$^*$~\cite{wen2023cap}      & \underline{23.62} & \underline{0.90} &  \underline{0.12}            \\
		Neural Preset~\cite{ke2023neural}    & 20.61 & 0.86          & 0.15          \\
		Neural Preset$^*$~\cite{ke2023neural} & 23.06 & 0.89          & \underline{0.12}          \\
		ICTone                                & \textbf{25.83} & \textbf{0.94} & \textbf{0.06} \\ \hline
	\end{tabular}
	\label{table:psnr_ssim_lpips}
\end{table}

\subsection{Additional Details about User Study}
\label{sec:suppl_user}
\begin{figure}[!tb]
	\centering
	\includegraphics[width=0.7\linewidth]{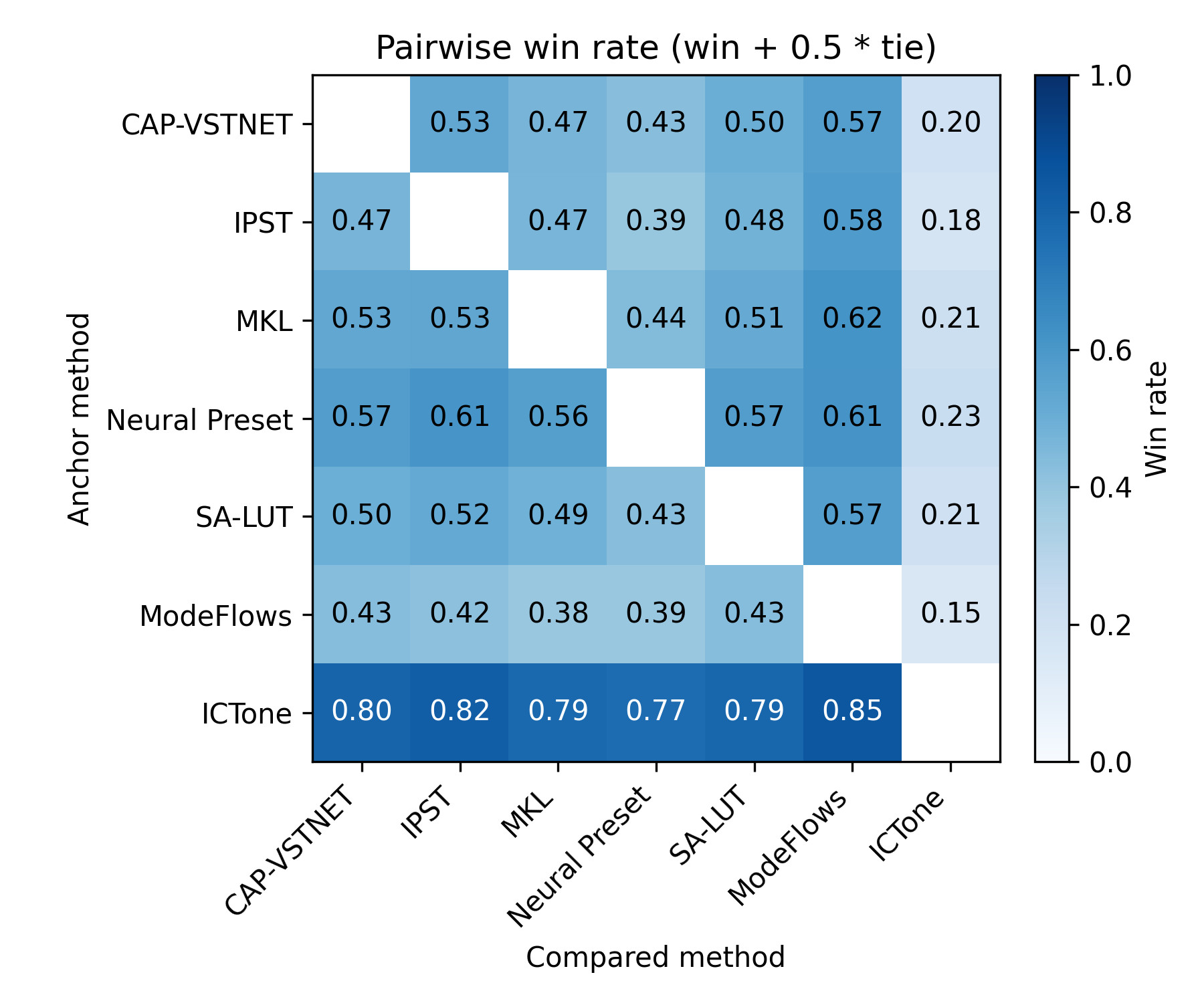}
	\vspace{-3mm}
    \caption{Pairwise win-rate matrix among the seven methods. Each entry indicates the percentage of times one method was preferred over another across all users and triplets. Darker colors correspond to higher win rates.}
    \vspace{-3mm}
	\label{fig:user_rank_paiwise_matrix}
\end{figure}
To complement the average ranking results reported in the main paper, we provide a more detailed analysis of the user study. Fig.~\ref{fig:user_rank_paiwise_matrix} presents a $7\times 7$ matrix where each entry indicates the percentage of times one method was preferred over another across all users and triplets. Darker colors correspond to higher win rates, offering a clearer view of relative preferences beyond mean rankings.
\subsection{Additional Visual Results}
\label{sec:suppl_visual}
\begin{figure*}[!tb]
	\centering
	\includegraphics[width=1.0\linewidth]{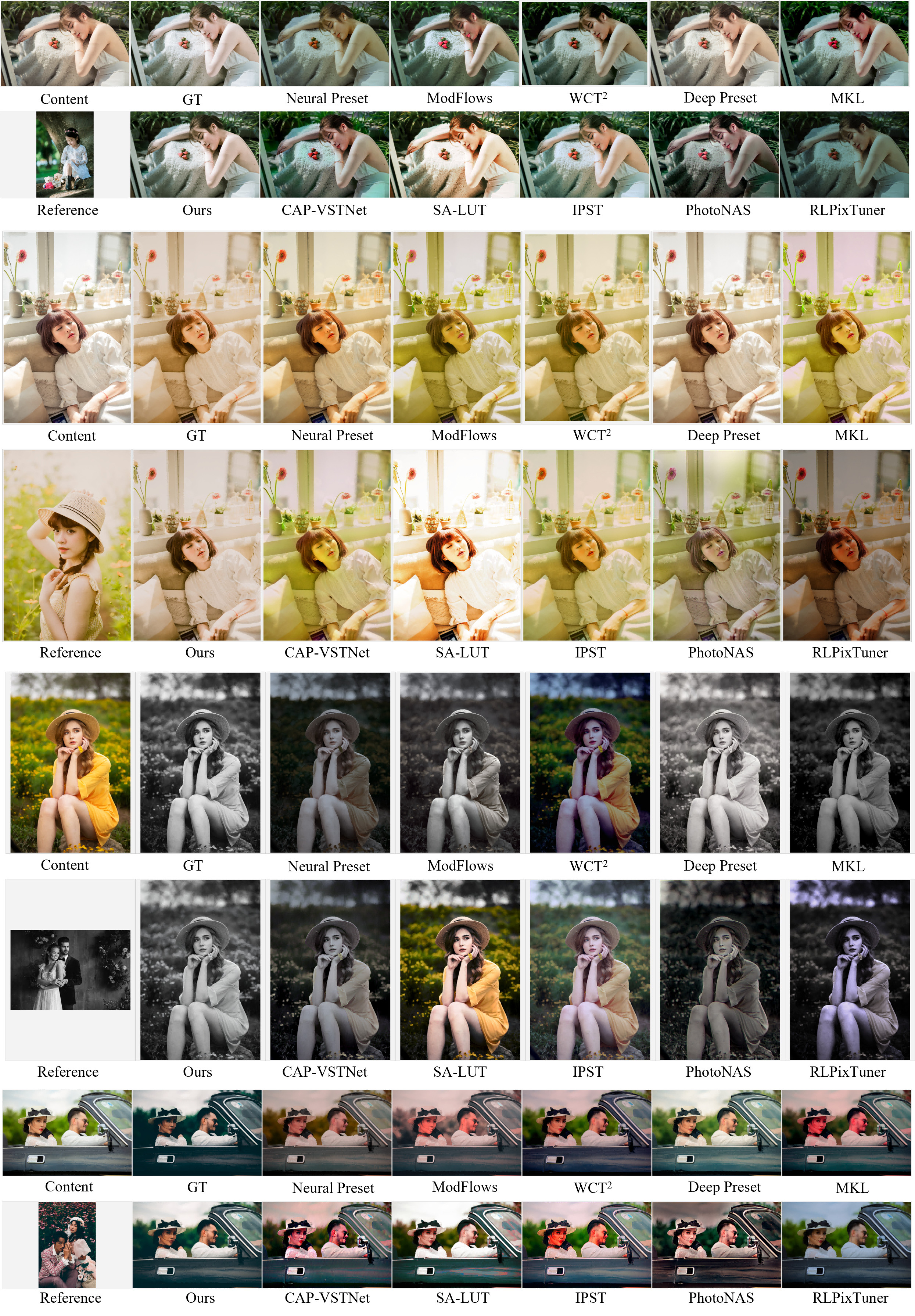}
	\caption{Additional Qualitative comparisons on TST2K (portrait scenes) including ground-truth (GT).
	Each group contains the content image, the reference image, our method (ICTone) result, and results of ten competing methods, with GT included to facilitate direct comparison. ICTone not only transfers the reference tone style effectively but also produces results closer to GT, with natural and consistent skin tones compared to other methods.}
	\label{fig:showcase_comvis1}
\end{figure*}

\begin{figure*}[!tb]
	\centering
	\includegraphics[width=1.0\linewidth]{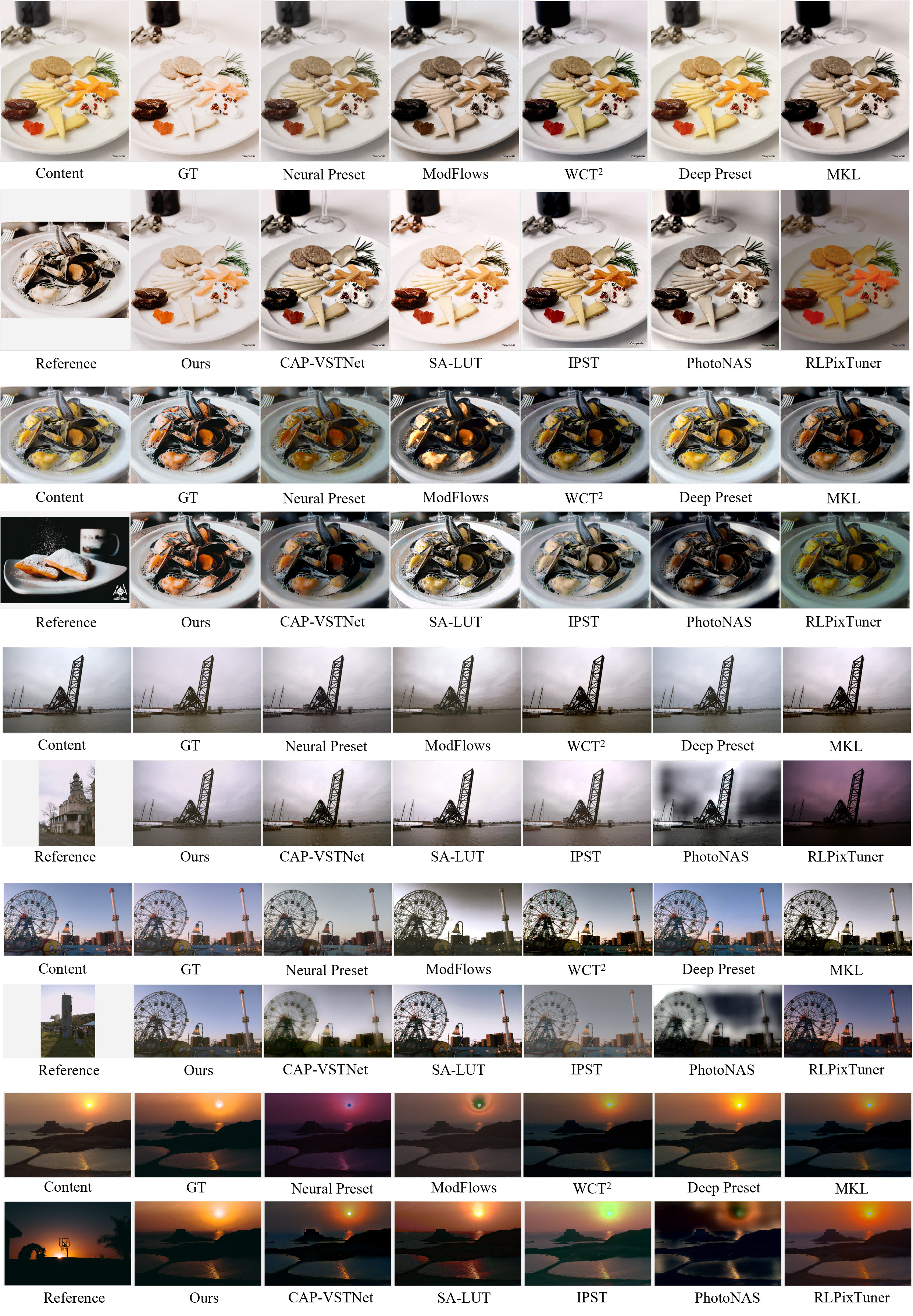}
	\caption{Additional Qualitative comparisons on TST2K (food, landscape and lifestyle) including ground-truth (GT).
		Each group contains the content image, the reference image, our method (ICTone) result, and results of ten competing methods, with GT included to facilitate direct comparison. ICTone aligns more closely with GT in tone style while avoiding artifacts such as color bleeding and unnatural saturation observed in other methods.}
	\label{fig:showcase_comvis2}
\end{figure*}

We provide extended comparisons against state-of-the-art methods, including ground-truth (GT) stylized images. As shown in Fig.~\ref{fig:showcase_comvis1} and Fig.~\ref{fig:showcase_comvis2}, the inclusion of GT offers a clearer perspective on the upper bound of performance, thereby enabling a more comprehensive evaluation of each method. 

These extended results demonstrate that while existing approaches achieve competitive performance in either content preservation or style similarity, they often struggle to balance both aspects simultaneously. In contrast, our proposed ICTone consistently achieves results that are closer to GT across multiple benchmarks, confirming its ability to maintain semantic fidelity while accurately transferring stylistic attributes. The comparisons with GT serve as a strong reference point, validating the robustness and generalization capability of our method.
\subsection{Extension to Colorization Tasks}
\label{sec:suppl_colorization_task}
\begin{figure}[!h]
	\centering
	\includegraphics[width=0.6\linewidth]{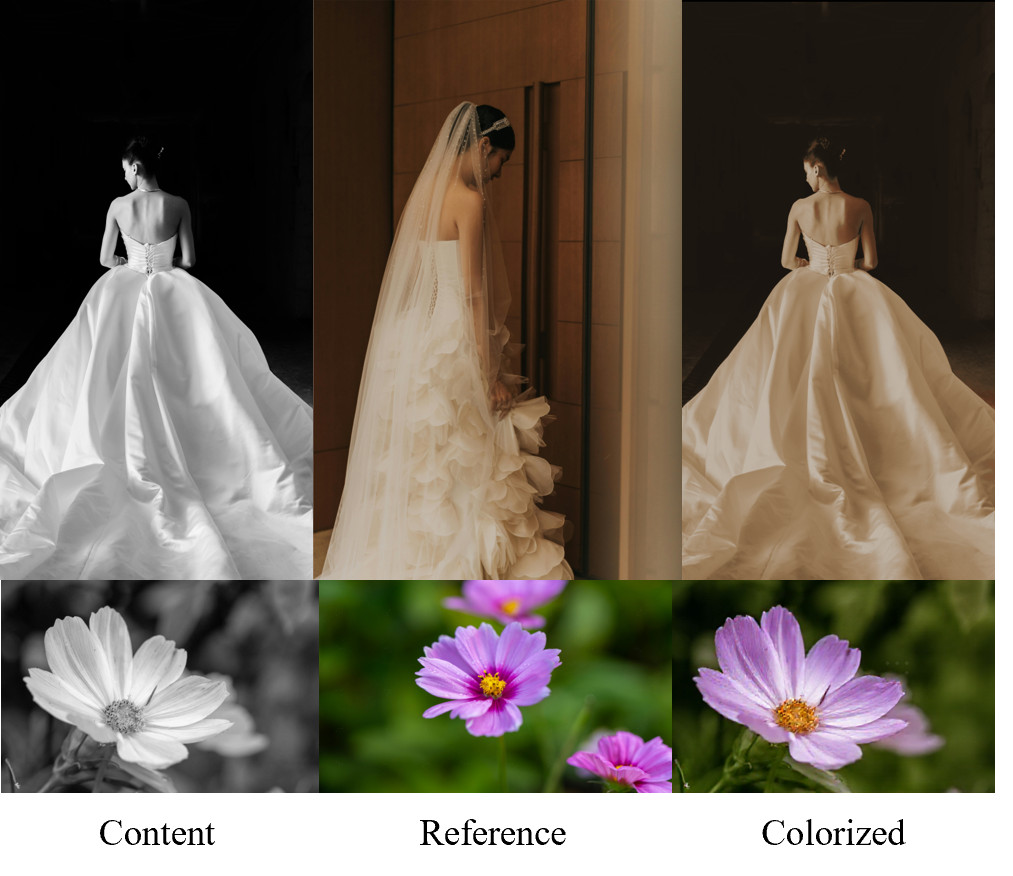}
    \vspace{-3pt}
	\caption{Extension to image colorization. From left to right: grayscale content image, reference image providing color cues, and the final colorized output. The result demonstrates that our framework preserves structural details while transferring realistic color distributions, highlighting its versatility across visual transformation tasks.}
    \vspace{-6pt}
	\label{fig:colorization_example}
\end{figure}

Beyond tone style transfer, our ICTone can be naturally extended to image colorization. In this setting, the input is a grayscale content image, and the model leverages reference images to guide the colorization process. As shown in Fig.~\ref{fig:colorization_example},  ICTone successfully restores natural color tones while maintaining the semantic structure of the grayscale input.
\section{Discussion and Limitations}
\label{sec:suppl_limit}
\subsection{Limitations}
The proposed dataset construction pipeline still has several limitations. 
During dataset construction, existing aesthetic assessment models are observed to exhibit stylistic biases and limited robustness. A fixed aesthetic score threshold may therefore remove images with valid but highly personalized tone styles. To mitigate this issue, an aesthetic non-degradation constraint is adopted, which retains stylized results whose aesthetic score is not lower than that of the original image. This strategy preserves overall visual quality while allowing diverse tone styles to remain in the dataset. Nevertheless, stronger aesthetic models could further improve the filtering process and retain more high-quality personalized samples.

Another limitation lies in the reliance on static preset enumeration when generating candidate stylized images. Although this strategy enables scalable data generation, it does not fully reflect the adaptive decision process used in professional photo retouching. A promising future direction is agentic dataset construction, where MLLM-based agents such as JarvisArt~\cite{jarvisart2025} analyze the content image and recommend suitable presets or invoke color adjustment tools when necessary. Such a pipeline could better simulate real retouching workflows and enable more realistic and content-adaptive triplet generation.

\clearpage

\end{document}